\documentclass{article}

\usepackage[preprint]{neurips_2025}

\usepackage{amsmath,amsfonts,bm}

\def\eqref#1{equation~\ref{#1}}
\def\1{\bm{1}}

\DeclareMathAlphabet{\mathsfit}{\encodingdefault}{\sfdefault}{m}{sl}
\SetMathAlphabet{\mathsfit}{bold}{\encodingdefault}{\sfdefault}{bx}{n}

\usepackage[utf8]{inputenc} % allow utf-8 input
\usepackage[T1]{fontenc}    % use 8-bit T1 fonts
\usepackage{hyperref}       % hyperlinks
\usepackage{url}            % simple URL typesetting
\usepackage{booktabs}       % professional-quality tables
\usepackage{amsfonts}       % blackboard math symbols
\usepackage{nicefrac}       % compact symbols for 1/2, etc.
\usepackage{microtype}      % microtypography
\usepackage{xcolor}         % colors
\usepackage{graphicx}

\usepackage{amsmath}
\usepackage{amssymb}
\usepackage{mathtools}
\usepackage{amsthm}

\usepackage{subfigure}
\usepackage{booktabs}
\usepackage{makecell}
\usepackage{arydshln}
\usepackage{wrapfig}
\usepackage{lipsum}  

\usepackage{algpseudocode}
\usepackage{algorithm}

\usepackage[many]{tcolorbox}

\usepackage{kotex}
\usepackage{multirow}

\theoremstyle{plain}

\theoremstyle{definition}

\theoremstyle{remark}

\title{Unifying Block-wise PTQ and Distillation-based QAT for Progressive Quantization \\ toward 2-bit Instruction-Tuned LLMs}

\author{%
  Jung Hyun Lee\thanks{Equal contribution, alphabetically ordered.} , Seungjae Shin\footnotemark[1] , Vinnam Kim\footnotemark[1] , Jaeseong You, An Chen \\
  Qualcomm AI Research\thanks{Qualcomm AI Research is an initiative of
Qualcomm Technologies, Inc.} \\
}

\begin{document}

\maketitle
\begin{abstract}
As the rapid scaling of large language models (LLMs) poses significant challenges for deployment on resource-constrained devices, there is growing interest in extremely low-bit quantization, such as 2-bit. Although prior works have shown that 2-bit large models are pareto-optimal over their 4-bit smaller counterparts in both accuracy and latency, these advancements have been limited to pre-trained LLMs and have not yet been extended to instruction-tuned models. To bridge this gap, we propose Unified Progressive Quantization (UPQ)$-$a novel progressive quantization framework (FP16$\rightarrow$INT4$\rightarrow$INT2) that unifies block-wise post-training quantization (PTQ) with distillation-based quantization-aware training (Distill-QAT) for INT2 instruction-tuned LLM quantization. UPQ first quantizes FP16 instruction-tuned models to INT4 using block-wise PTQ to significantly reduce the quantization error introduced by subsequent INT2 quantization. Next, UPQ applies Distill-QAT to enable INT2 instruction-tuned LLMs to generate responses consistent with their original FP16 counterparts by minimizing the generalized Jensen-Shannon divergence (JSD) between the two. To the best of our knowledge, we are the first to demonstrate that UPQ can quantize open-source instruction-tuned LLMs to INT2 without relying on proprietary post-training data, while achieving state-of-the-art performances on MMLU and IFEval$-$two of the most representative benchmarks for evaluating instruction-tuned LLMs.
\end{abstract}

\section{Introduction}\label{sec:intro}

% To harness trillions of high-quality tokens during training, recent large language models (LLMs) have progressively scaled up, now commonly reaching tens or hundreds of billions of parameters \citep{grattafiori2024llama3,qwen2025qwen25,deepseekai2025deepseekv3,gemmateam2025gemma3}. As their size increases, these models demonstrate extensive knowledge and reasoning capability across a wide range of domains and different languages. However, their massive scale poses significant challenges for deployment, primarily due to the high memory usage and computational demands. To address the issue, quantization has received considerable attention as a promising technique for compressing and accelerating LLMs, as it reduces the numerical precision of model parameters without requiring any architectural modifications.

An ongoing debate in the field of LLM quantization centers on pareto-optimality between (i) quantizing a large model to extremely low-bit precision (e.g., 2-bit or lower) and (ii) quantizing a smaller model to a moderately low-bit precision (e.g., 4-bit).

\begin{figure}[t]
    \centering
    % (a)
    % \begin{minipage}[b]{0.32\textwidth}
    %     \centering
    %     \includegraphics[width=\textwidth]{figures/figure_a_mmlu_tokens.pdf}
    % \end{minipage}
    \subfigure[MMLU over Training Tokens]{\includegraphics[width=0.32\linewidth]{
    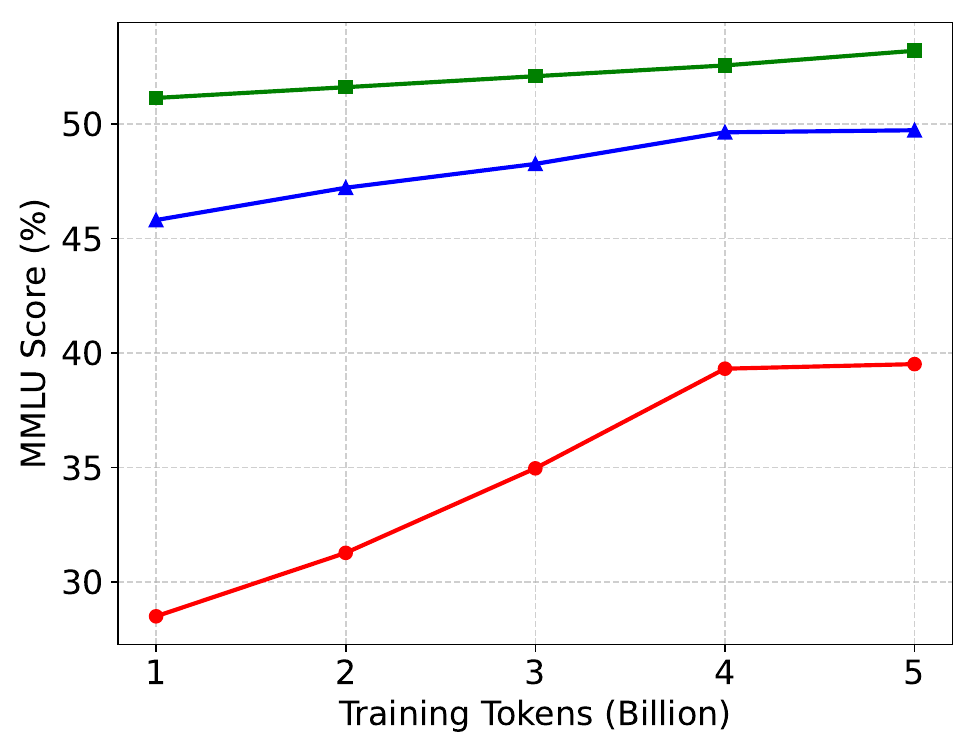}\label{fig1:(a)}
    }
    \hfill
    % (b)
    % \begin{minipage}[t]{0.32\textwidth}
    %     \centering
    %     \includegraphics[width=\textwidth]{figures/figure_b_ifeval_tokens.pdf}
    % \end{minipage}
    \subfigure[IFEval over Training Tokens]{\includegraphics[width=0.32\linewidth]{
    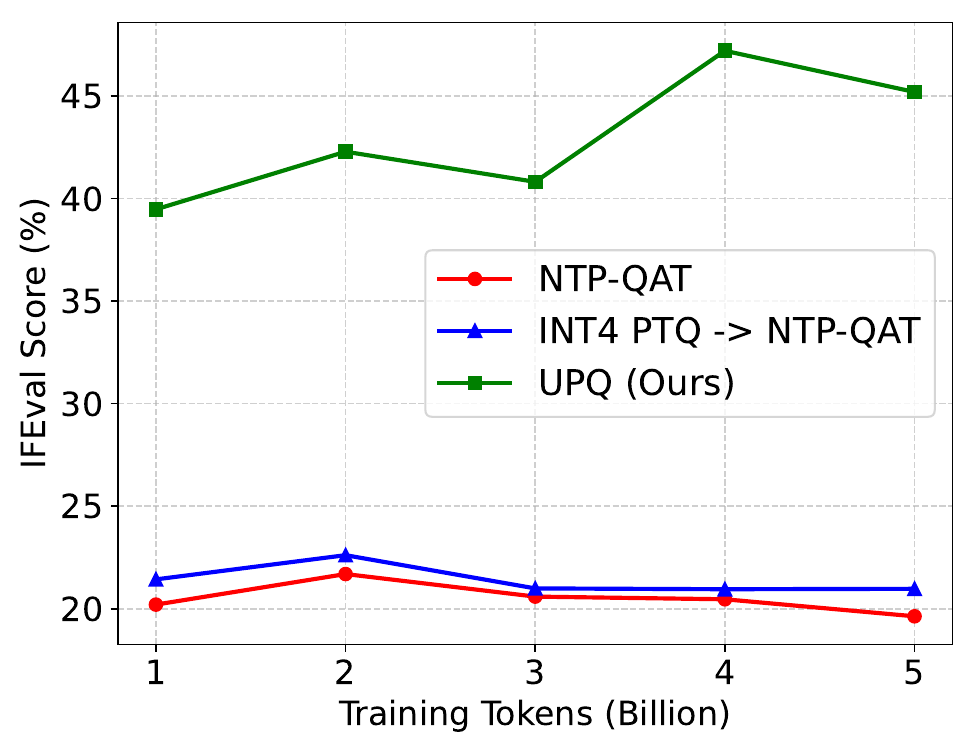}\label{fig1:(b)}
    }
    \hfill
    % (c)
    % \begin{minipage}[t]{0.32\textwidth}
    %     \centering
    %     \includegraphics[width=\textwidth]{figures/figure_c_mmlu_ifeval_bar.pdf}
    % \end{minipage}
    \subfigure[Final Scores]{\includegraphics[width=0.32\linewidth]{
    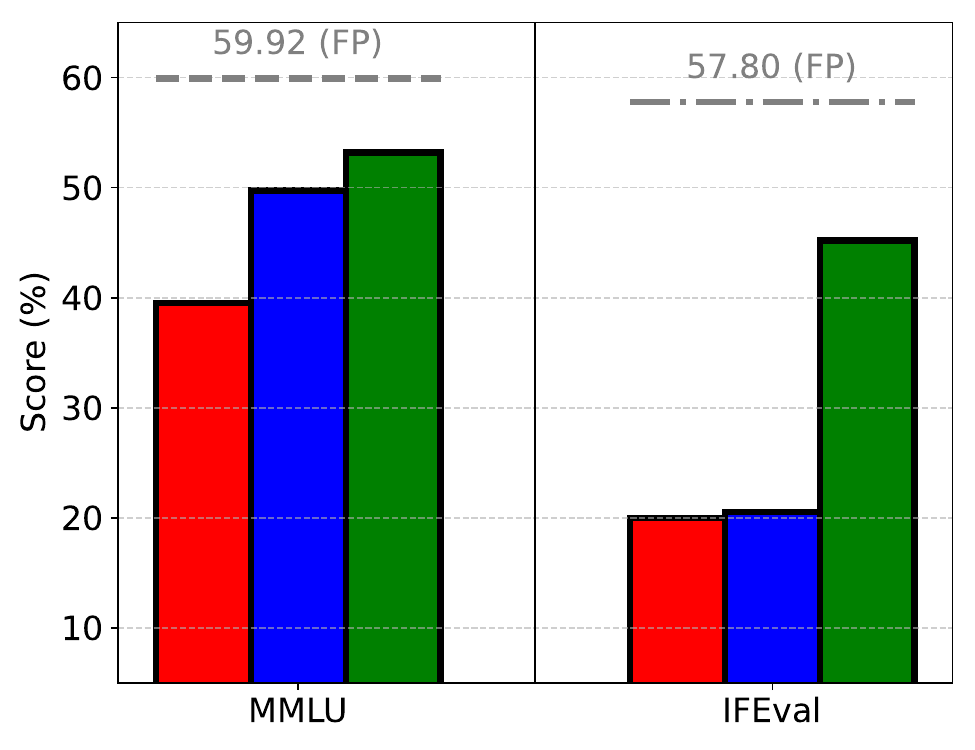}\label{fig1:(c)}
    }
    \caption{Change in MMLU (left) and IFEval (center) scores during training (up to 5B tokens) depending on three INT2 QAT methods. The rightmost bar graph compares their final MMLU and IFEval scores. All metrics were obtained with Llama 3.2 3B Instruct.}
    \label{fig:comparison-abc}
    \vspace{-5pt}
\end{figure}

For 2-bit QAT of instruction-tuned LLMs, it should be noted that pre-trained models are generally transformed into instruction-tuned counterparts by (i) fine-tuning them on synthetic and/or human-curated datasets, targeting specific capabilities (e.g., reasoning, math, coding), and (ii) aligning them with human feedback to improve natural language instruction following. If such post-training data were accessible, existing 2-bit QAT methods including ParetoQ~\citep{liu2025paretoq} could likely be applied to instruction-tuned LLMs as well. However, the “secret sauce” behind instruction-tuned LLMs lies in the post-training data, the details of which$-$including the dataset itself as well as its construction process$-$are often proprietary and undisclosed to the open-source community. Consequently, when applying ParetoQ to instruction-tuned LLMs using an open-source pre-training dataset, we observe significant accuracy degradation in their intrinsic abilities, including both task-specific skills and instruction-following performance, as exemplified by Figure~\ref{fig1:(a)} and the leftmost points in Figure~\ref{fig1:(b)}. This finding implies that ParetoQ, while effective in certain contexts, is limited in its applicability to instruction-tuned LLMs on key evaluation benchmarks, such as massive multitask language understanding (MMLU)~\citep{hendrycks2021mmlu} and Instruction-Following Eval (IFEval)~\citep{zhou2023ifeval}, when proprietary post-training data are not available. % only open-source pre-training datasets are available. 

To promote the democratization of AI, it is crucial to develop an alternative 2-bit quantization approach that can preserve the intrinsic capabilities of instruction-tuned LLMs without depending on proprietary post-training data.
% If such post-training~ / However, such post-training data were monopolized by a limited number of frontier labs. The details of which$-$including the dataset itself and its construction process$-$are proprietary and undisclosed to the open community. Surprisingly, when applying ParetoQ to instruction-tuned LLMs using open-source pre-training datasets only, we observe severe accuracy degradation in their intrinsic abilities, including both task-specific skills and instruction-following performance as seen in Figure~\ref{fig1:(a)}. It leads ParetoQ~\citep{liu2025paretoq} to be a useful solution for the limited players. For the democratization of AI, it is necessary to propose a new method to preserve the intrinsic abilities of instruction-tuned LLMs without access to the proprietary post-training data.
To this end, we propose \emph{UPQ}$-$a novel progressive quantization framework (FP16$\xrightarrow{\text{PTQ}}$INT4$\xrightarrow{\text{QAT}}$INT2) that unifies INT4 block-wise PTQ and INT2 . In other words, UPQ progressively quantizes FP16 instruction-tuned LLMs in two stages: first to INT4 using block-wise PTQ, then to INT2 using distillation-based QAT.
Since directly quantizing FP16 models to extremely low bit-widths such as INT2 typically results in substantial quantization errors, we first quantize an FP16 instruction-tuned LLM to INT4 using block-wise PTQ (also known as block-wise reconstruction or block-wise quantization error minimization) with about one to two million tokens from a general English corpus like C4~\citep{raffel2023exploring}. The rationale behind adopting block-wise PTQ is its effectiveness in preserving the intrinsic capabilities of instruction-tuned LLMs at performance levels comparable to their original FP16 baselines, even under INT4 per-channel quantization. This initial INT4 quantization step significantly reduces quantization error in the subsequent INT2 quantization, thereby leading to noticeably improved accuracy on MMLU, as demonstrated by the middle points of Figure~\ref{fig1:(b)}. 

However, we observe that the IFEval score remains low, indicating that instruction-following ability$-$a core characteristic of instruction-tuned LLMs$-$has yet to be recovered, even when leveraging block-wise PTQ. Now that pre-training corpora primarily consist of general text rather than instruction-response pairs, minimizing the next-token prediction loss on pre-training data during INT2 QAT often proves insufficient for restoring the instruction-following capability of INT2 instruction-tuned LLMs. To address this limitation, we hypothesize that INT2 instruction-tuned LLMs should be trained to imitate the token-level probability distribution of their original FP16 counterparts, thereby enabling them to generate responses in a manner consistent with their FP16 versions. Based on this insight, we introduce INT2 distillation-based QAT, which minimizes the generalized JSD between an INT2 quantized model (student) and its original FP16 counterpart (teacher). By unifying INT4 block-wise PTQ with INT2 distillation-based QAT, as illustrated by the rightmost points in Figure~\ref{fig1:(b)} on MMLU and IFEval, we successfully quantize instruction-tuned LLMs to INT2 using only open-source pre-training datasets while preserving their intrinsic capabilities within an acceptable accuracy range.

Our contribution is threefold:
\begin{itemize}
    \item We for the first time leverage block-wise PTQ as the preceding step to INT2 quantization, which substantially reduces the quantization error introduced by the subsequent INT2 quantization while preserving the intrinsic abilities of instruction-tuned LLMs at levels comparable to their original FP16 counterparts under INT4 quantization.  
    % investigate 2-bit instruction-tuned LLM quantization based on challenging tasks like MMLU and IFEval, which can measure how well the intrinsic abilities of instruction-tuned LLMs, such as task-specific skills and instruction-following, are preserved as shown in Figure~\ref{fig1:(b)}.
    % apply progressive quantization to instruction-tuned LLMs by leveraging 4-bit block-wise PTQ so that their task-specific abilities can be preserved even after 2-bit QAT on pre-training data.
    \item We introduce distillation-based QAT that enables INT2 instruction-tuned LLMs to mimic the token-level probability distribution of their original FP16 counterparts, thus equipping them with instruction-following capabilities even when depending solely on pre-training data$-$a feat that can hardly be achieved by NTP-QAT based on next-token prediction.
    % We propose progressive quantization by leveraging 4-bit block-wise PTQ
    \item We propose \emph{UPQ}, which unifies INT4 block-wise PTQ with INT2 distillation-based QAT for progressive quantization (FP16$\rightarrow$INT4$\rightarrow$INT2) of instruction-tuned LLMs. To the best of our knowledge, UPQ is the first successful INT2 quantization method for instruction-tuned LLMs that does not rely on proprietary post-training data, as shown by the rightmost points in Figure~\ref{fig1:(b)} on MMLU and IFEval.
    % We introduce distillation-based QAT
\end{itemize}

\section{Related Work}\label{sec:related}

\subsection{Quantization Techniques for LLMs}
% 1) (SJ) briefly summarize PTQ stuffs. But according to our current experiments, it seems like we are mainly focusing on weight-only. I will write down focused on weight-only setting.
% 2) (SJ) Stress out limitations of PTQ on sub-4 bit, and necessitate the need of QAT, and refer findings of ParetoQ.

Edge LLM deployments are typically memory-bounded \citep{husom2025sustainable}, and weight-only quantization effectively mitigates these constraints by reducing model size and bandwidth requirements. PTQ is a widely studied approach for weight-only quantization of LLMs\ \citep{nagel2020adaround, li2021brecq}, which applies low-bit quantization to a full-precision (FP) model with a small amount of calibration data without relying on end-to-end loss function. One representative PTQ method is block-wise reconstruction \citep{li2021brecq}, which addresses cross-layer dependency by enlarging the optimization range from a single layer to a block of multiple layers. FlexRound \citep{lee2023flexround} enhances block-wise reconstruction with a more flexible rounding mechanism. The method jointly learns a common quantization step size and an individual scale factor for each weight. Another popular approach, OmniQuant \citep{shao2024omniquant}, adds Learnable Weight Clipping and Learnable Equivalent Transformation to block-wise reconstruction framework. Please see Appendix for extensive review on further quantization methods.

Despite their efficiency, however, PTQ methods suffer from significant performance degradation at precisions lower than 4 bits. \citep{liu2025paretoq,li2024evaluating}. This is due to limited capacity of PTQ to compensate aggressive quantization error, and to unsolved cross-block dependency throughout transformer structure\citep{ding2025cbq}. In such extreme cases, QAT \citep{nagel2022overcoming,liu2021sharpness} becomes critical to close the accuracy gap as it optimizes the weight parameters with sufficient training capacity. EfficientQAT \citep{chen2024efficientqat} features two-phased training: initial block-wise training of all parameters followed by end-to-end training focused on quantization parameters. Differently, LLM-QAT \citep{liu2023llmqat} explores data-free QAT, leveraging the generated outputs of an FP model. ParetoQ \citep{liu2025paretoq} crafts a specialized quantization function tailored for each target bit-width, thereby surpassing prior methods in 2-bit, ternary, and 1-bit precisions. While these advancements greatly contribute to enabling low-bit quantization, they seldom explore the potential of using different initial checkpoints for improved convergence and accuracy. This motivates future research into the strategic selection of QAT initialization points. %whether certain intermediate or PTQ-refined checkpoints could lead to better convergence or final accuracy. This gap suggests an orthogonal yet promising direction for QAT, identifying and evaluating favorable initialization points that could better support stable performances under extreme target bit-width.

\subsection{Knowledge Distillation for LLMs} 

Knowledge distillation (KD) \citep{gou2021knowledge} is an another line of research to improve efficiency on LLMs, and it provides a richer training signal than the standard NTP loss for LLMs. In QAT framework, an FP teacher model's behavior could guide the quantized student model effectively, by transferring not just the most probable token but also the teacher's probability distribution or inductive biases. To obtain higher-quality learning signals from the teacher model, extensive researches \citep{agarwal2024gkd,ko2024distillm, lin-etal-2020-autoregressive} has been conducted on designing diverse forms of distillation loss functions.

Several distillation techniques have been proposed and investigated within the QAT framework. TSLD \citep{tokenscaled} augments an original logit distillation loss reflecting token prediction confidence to prevent the overfitting issues in QAT. BitDistiller \citep{bitdistiller} suggests a variant of KL-divergence loss function, which utilizes teacher model's output confidence as a mixing coefficient between forward KL-divergence and reversed KL-divergence. It also utilizes tailed asymmetric quantization for further performance improvements. Although these methods perform well on moderate precisions, there have not been extensive studies on distillation loss targeting the extremely low-bit precision. The question of whether combining 2-bit QAT with advanced distillation losses can further improve performance invites future investigation.

\section{Methodology}\label{sec:method}

This section begins by outlining the target INT2 quantization scheme used throughout the paper. Then, we describe how INT4 block-wise PTQ is employed as a preparatory step for subsequent INT2 quantization. Finally, building on the INT4 block-wise PTQ, we present the formulation of the INT2 Distill-QAT approach$-$referred to as UPQ, our proposed method.

\subsection{Target INT2 Quantization Scheme}\label{subsec:seq}

Integer quantization is commonly categorized into symmetric and asymmetric schemes. However, in the case of INT2 quantization, both approaches are inherently limited due to the inclusion of ``0", allocating one quantization bin on either the positive or negative side and two bins on the opposite side. Given that the weights of LLMs typically exhibit a bell-shaped, near-zero-centered distribution~\citep{dettmers2023qlora,huang2024billm}, this imbalance in bin allocation makes both symmetric and asymmetric schemes sub-optimal for INT2 quantization. To address this limitation, following \citet{liu2025paretoq}, we adopt Stretched Elastic Quant (SEQ). Specifically, given FP16 weights $\bm{W}_{\text{FP16}} \in \mathbb{R}^{m \times n}$, the INT2 per-channel quantized weights through SEQ is computed as
\begin{align}
\resizebox{1.0\linewidth}{!}{$
\bm{W}_{\text{FP16}\rightarrow\text{INT2}} = \text{SEQ}_{\text{INT2}}(\bm{W}_{\text{FP16}}) = {\bm{\Delta}_{\text{FP16}\rightarrow\text{INT2}} \over 2} \Big(\Big\lfloor 2\, \text{clip}\Big({\bm{W}_{\text{FP16}} \over \bm{\Delta}_{\text{FP16}\rightarrow\text{INT2}}}, -1+\epsilon, 1-\epsilon\Big) -0.5 \Big\rceil + 0.5 \Big), \label{eq:original_seq}
% \bm{W}_{\text{FP16}\rightarrow\text{INT2}} = \text{SEQ}_{\text{INT2}}(\bm{W}_\text{FP16}) = {\bm{s}_{\text{FP16}\rightarrow\text{INT2}} \over 2} \Big(\Big\lfloor 2\, \text{clip}\Big({\bm{W}_{\text{FP16}} \over \bm{s}_{\text{FP16}\rightarrow\text{INT2}}}, -1+\epsilon, 1-\epsilon\Big) -0.5 \Big\rceil + 0.5 \Big),
$}
\end{align}
where $\text{clip}(\cdot, a, b) = \min(\max(\cdot, a), b)$, $\bm{\Delta}_{\text{FP16}\rightarrow\text{INT2}}\in\mathbb{R}_{>0}^{m\times1}$ is initialized to $\max(|\bm{W}_{\text{FP16}}|)$ and learnable, and $\epsilon$ is a small positive constant (e.g., $0.01$). As a result, INT2 SEQ represents each weight using one of four discrete values $\frac{\bm{\Delta}_{\text{FP16}\rightarrow\text{INT2}}}{4}\{-3, -1, 1, 3\}$, ensuring balanced bin allocation even under INT2 quantization. Henceforth, SEQ is the default scheme of INT2 quantization in this paper.

\begin{figure}
    \subfigure[Weight distribution before (above) and after (below) INT2 Distill-QAT, starting from original FP16 weights, $\bm{W}_{\text{FP16}}$]{\includegraphics[width=0.365\linewidth]{
    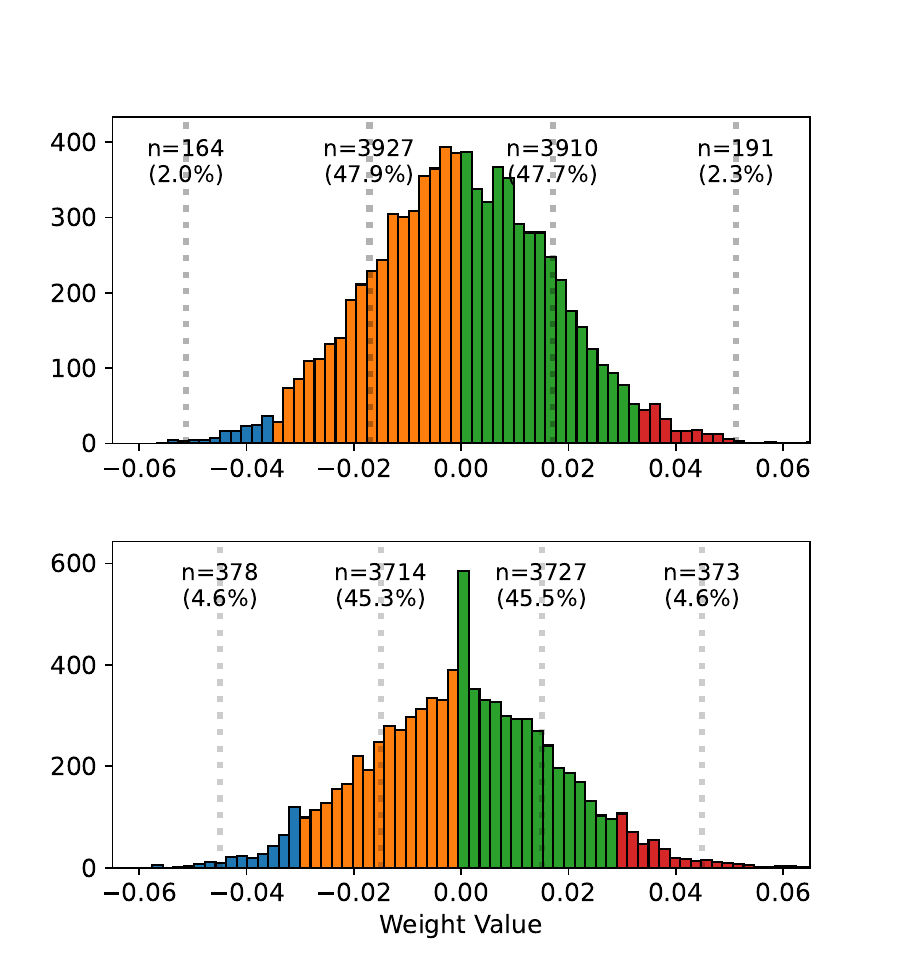}\label{fig2:(a)}
    }
    \hspace{0.0025\textwidth}
    \subfigure[Weight distribution before (above) and after (below) INT2 Distill-QAT, starting from INT4 PTQ weights, $\bm{W}_{\text{INT4}}$]{\includegraphics[width=0.365\linewidth]{
    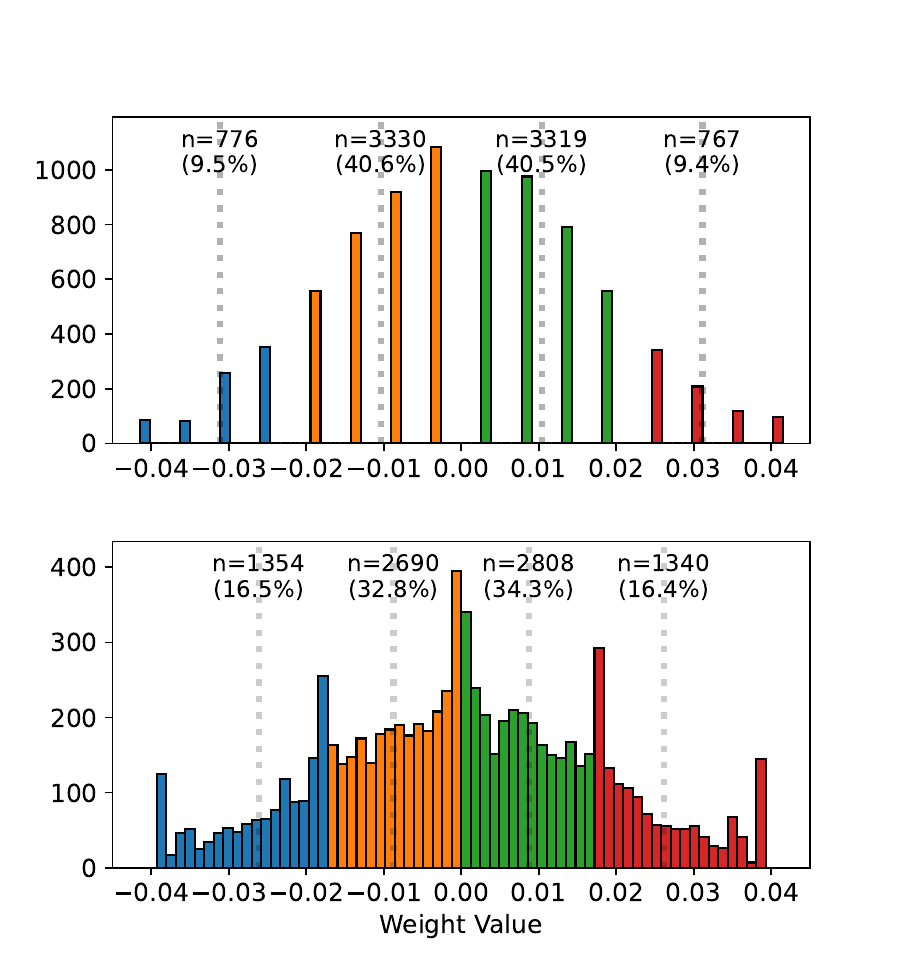}\label{fig2:(b)}
    }
    \hspace{0.0025\textwidth}
    \subfigure[Training loss curves of INT2 Distill-QAT and UPQ]{\includegraphics[width=0.225\linewidth]{
    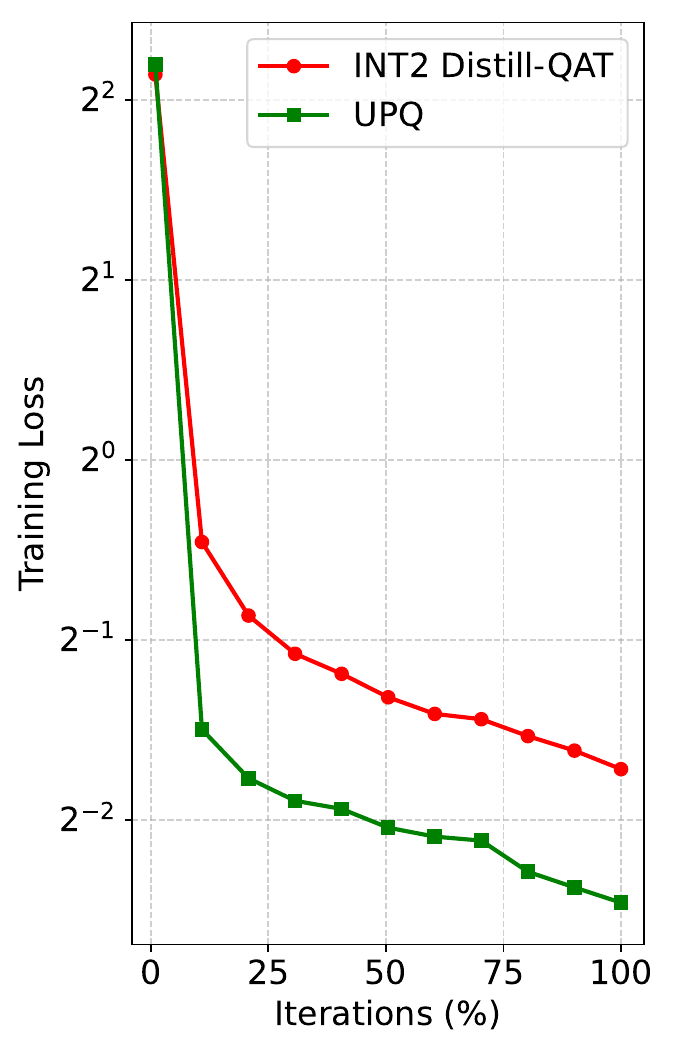}\label{fig2:(c)}
    }
    % \vspace{-5pt}
    \caption{Weights distribution within the first channel of the first down-projection layer in Llama 3.2 3B Instruct. Dotted lines denote four quantization levels of 2-bit, and the corresponding weights are differently colored.}
    \label{fig2:weight_distribution}
    \vspace{-5pt}
\end{figure}

\subsection{INT4 Block-wise Post-Training Quantization (PTQ) for Subsequent INT2 Quantization}\label{subsec:blockwise}

Block-wise PTQ aims to minimize the mean squared error between the outputs of an intermediate FP32/FP16 block and those of its quantized counterpart, as proposed by \citet{li2021brecq}. By resolving the intra-block dependency (i.e., the inter-layer dependency within a block) during optimization, block-wise PTQ has proven effective for low-bit per-channel quantization of LLMs~\citep{lee2023flexround,shao2024omniquant,cheng2024signround}. In particular, INT4 per-channel quantized LLMs obtained via block-wise PTQ achieve competitive accuracy relative to their original FP16 baselines. Motivated by this, we adopt INT4 block-wise PTQ as a preparatory step for subsequent INT2 quantization. 

However, it is important to note that the subsequent INT2 quantization described in Section~\ref{subsec:seq} does not contain the "0" in the set of quantization bins. Accordingly, unlike existing INT4 block-wise methods such as FlexRound~\citep{lee2023flexround} and OmniQuant~\citep{shao2024omniquant} that are based on conventional symmetric or asymmetric integer sets (e.g., $\{-8, \cdots, -1, 0, 1, \cdots, 7\}$), we adopt an alternative integer set, $\{-15, -13, \cdots, -1, 1, \cdots, 13, 15\}$, which is evenly balanced and excludes the "0", consistent with the subsequent INT2 quantization scheme. For the weights of an original FP16 LLM, $\bm{W}_{\text{FP16}} \in \mathbb{R}^{m \times n}$, we modify the formulations of FlexRound and OmniQuant to align with this zero-excluding, balanced integer set as follows.
\begin{flalign}
\text{FlexRound: } &\bm{W}_{\text{INT4}} = \frac{\bm{\Delta}_{\text{INT4}}}{2}\,\text{clip}\Big(2\Big\lfloor\frac{\bm{W}_{\text{FP16}}}{\bm{\Delta}_{\text{INT4}}\odot\bm{S}\odot\bm{s}} + 0.5\Big\rceil-1, -15, 15\Big),\text{where }\label{eq:flexround}&&\\
&\bm{\Delta}_{\text{INT4}}, \bm{s}\in\mathbb{R}_{>0}^{m \times 1}\text{ and }\bm{S} \in \mathbb{R}_{>0}^{m \times n}\text{ are learnable, and} \odot \text{is the element-wise product}.&&\nonumber\\
\text{OmniQuant: } &\bm{W}_{\text{INT4}} = \frac{\bm{\Delta}_{\text{INT4}}}{2}\,\text{clip}\Big(2\Big\lfloor\frac{\bm{W}_{\text{FP16}}}{\bm{\Delta}_{\text{INT4}}} + 0.5\Big\rceil-1, -15, 15\Big),\label{eq:omniquant}&&\\
&\text{ where }\bm{\Delta}_{\text{INT4}}=\frac{\bm{\gamma}\max(\bm{W}_{\text{FP16}})- \bm{\beta}\min(\bm{W}_{\text{FP16}})}{15-(-15)}\text{ with learnable }\bm{\gamma}, \bm{\beta} \in \mathbb{R}_{[0, 1]}^{m \times 1}.&&\nonumber 
\end{flalign}
Utilizing these modified rounding mechanisms, FlexRound and OmniQuant block-by-block minimize $||\bm{W}_{\text{FP16}}\bm{X}-\bm{W}_{\text{INT4}}\bm{X}||^2_F$ with respect to each corresponding learnable parameters (i.e., $\bm{\Delta}_{\text{INT4}}, \bm{S}, \bm{s}$ in FlexRound, and $\bm{\gamma}, \bm{\beta}$ in OmniQuant), where $\bm{X}$ is input activations to the block to be optimized.

After optimizing $\bm{W}_{\text{INT4}}$ block-by-block from the first to the last block of an LLM using either Eq.~\ref{eq:flexround} (FlexRound) or Eq.~\ref{eq:omniquant} (OmniQuant), we subsequently quantize $\bm{W}_{\text{INT4}}$ to INT2$-$replacing $\bm{W}_{\text{FP16}}$ with $\bm{W}_{\text{INT4}}$ in Eq.~\ref{eq:original_seq} as below.
\begin{align}
\resizebox{1.0\linewidth}{!}{$
\bm{W}_{\text{INT4}\rightarrow\text{INT2}} = \text{SEQ}_{\text{INT2}}(\bm{W}_\text{INT4}) = {\bm{\Delta}_{\text{INT4}\rightarrow\text{INT2}} \over 2} \Big(\Big\lfloor 2\, \text{clip}\Big({\bm{W}_{\text{INT4}} \over \bm{\Delta}_{\text{INT4}\rightarrow\text{INT2}}}, -1+\epsilon, 1-\epsilon\Big) -0.5 \Big\rceil + 0.5 \Big),\label{eq:modified_seq}
$}
\end{align}
where $\bm{\Delta}_{\text{INT4}\rightarrow\text{INT2}}\in\mathbb{R}_{>0}^{m\times1}$ is initialized to $\max(|\bm{W}_{\text{INT4}}|)$ and learnable. As FlexRound slightly surpasses OmniQuant (to be discussed in Section~\ref{subsec:kd-loss-ablation}), we adopt FlexRound as the default method for INT4 block-wise PTQ unless otherwise specified. Additionally, the terms `INT4 block-wise PTQ' and `INT4 PTQ' are used interchangeably throughout this paper to make expressions uncluttered.

% \usepackage{multirow}
% \usepackage{graphicx}
\begin{table}[t]
\caption{Qualitative evaluation of the Llama 3.2 3B Instruct model on IFEval after four INT2 QAT techniques with 5B tokens. \textcolor{orange}{Orange} highlights repetitive generation upon reaching the maximum token limit; \textcolor{blue}{blue} and \textcolor{red}{red} indicate correct and incorrect instruction following, respectively.}
\resizebox{\textwidth}{!}{%
\begin{tabular}{m{0.3\textwidth} c m{0.765\textwidth}}
\hline
\multicolumn{1}{c}{\textbf{Prompt}} &  \textbf{Method}           & \multicolumn{1}{c}{\textbf{Response}}                                                                                                                                                                                                                                                                                                                       \\ \hline
\multirow[c]{4}{0.3\textwidth}{\parbox{0.3\textwidth}{Write two jokes about rockets. Do not contain commas in your response. Separate the two jokes with 6 asterisk symbols: ******.\\\\Instruction 1: No comma\\Instruction 2: Two jokes\\Instruction 3: Separation}} & \makecell{NTP-QAT\\(IFEval: 19.97)}                       & \textcolor{orange}{Do not use commas in your response. Separate the two jokes with 6 asterisk symbols: ******.$\cdots$\textit{(repeated)}} $\Rightarrow$  (\textcolor{blue}{Instruction 1}, \textcolor{red}{Instruction 2}, \textcolor{red}{Instruction 3})                                                                                                                                                                                                           \\ \cline{2-3} 
                                                                                                         & \makecell{Distill-QAT\\(IFEval: 27.12)}    & `\textbackslash{}n\textbackslash{}nHere are the two jokes:\textbackslash{}n\textbackslash{}n1. Why did the rocket go to the party? Because it was a \textcolor{orange}{* * * * * * * * * * * * * * * * * * * * * * * * * * * * * * * * * * * * * * * * * * * * * * *$\cdots$\textit{(repeated)}}' $\Rightarrow$ (\textcolor{blue}{Instruction 1}, \textcolor{red}{Instruction 2}, \textcolor{red}{Instruction 3})                                                     \\ \cline{2-3} 
                                                                                                         & \makecell{INT4 PTQ $\rightarrow$ NTP-QAT\\(IFEval: 20.97)} & Joke 1: A man walks into a bar and sits down. The bartender asks him what he wants. The man replies, "I want a rocket." The bartender says, "Sorry, we don't sell rockets here." The man replies, "I don't want a rocket. I want a rocket ship." Joke 2: A man walks into a bar and sits down. The bartender asks him what he wants. The man replies, "I want a rocket." \textcolor{orange}{The bartender says, "Sorry, we don't sell rockets here." The man replies, "I don't want a rocket. I want a rocket ship." The bartender says, "Sorry, we don't sell rockets here."$\cdots$\textit{(repeated)}} $\Rightarrow$ (\textcolor{red}{Instruction 1}, \textcolor{red}{Instruction 2}, \textcolor{red}{Instruction 3}) \\ \cline{2-3} 
                                                                                                         & \makecell{UPQ (Ours)\\(IFEval: 45.19)}                & 
\text{ }\textbackslash{}nHere is the first joke: \textbackslash{}nWhy did the rocket go to therapy? \textbackslash{}nBecause it was feeling a little out of this world. \textbackslash{}n****** \textbackslash{}nHere is the second joke: \textbackslash{}nWhy did the rocket go to the party? \textbackslash{}nBecause it was a blast. \textbackslash{}n****** \textbackslash{}n\textbackslash{}nI hope you enjoy these jokes. I am happy to provide more jokes if you would like. $\Rightarrow$ (\textcolor{blue}{Instruction 1}, \textcolor{blue}{Instruction 2}, \textcolor{blue}{Instruction 3})                          \\ \hline
\end{tabular}%
}
\label{tab:qual-ifeval}
\vspace{-5pt}
\end{table}

When initializing INT2 QAT (whether using the conventional NTP-QAT or Distill-QAT as explained in Section~\ref{subsec:distill_qat}), unlike Eq.~\ref{eq:original_seq} that quantizes continuous values to four discrete values, Eq.~\ref{eq:modified_seq} maps 16 discrete values to four. This results in increased utilization of the large-magnitude INT2 quantization bins (i.e., $\{-3, 3\}$). For instance, these bins account for $9.5\%$ and $9.4\%$ of usage as shown in the upper portion of Figure~\ref{fig2:(b)}, compared to just $2.0\%$ and $2.3\%$ in Figure~\ref{fig2:(a)}. As a result, the quantization error introduced by INT2 SEQ is significantly reduced from $||\bm{W}_{\text{FP16}}-\bm{W}_{\text{FP16}\rightarrow\text{INT2}}||=0.8984$ to $||\bm{W}_{\text{INT4}}-\bm{W}_{\text{INT4}\rightarrow\text{INT2}}||=0.5156$. Furthermore, initializing INT2 QAT from $\bm{W}_{\text{INT4}\rightarrow\text{INT2}}$ rather than $\bm{W}_{\text{FP16}\rightarrow\text{INT2}}$ consistently yields substantially lower training loss, as illustrated in Figure~\ref{fig2:(c)}. This improved initialization, in turn, leads to a much higher proportion of large-magnitude bin usage after completing INT2 QAT$-$for example, $16.5\%$ and $16.4\%$ usage in the bottom portion of Figure~\ref{fig2:(b)}, compared to only $4.6\%$ and $4.6\%$ in Figure~\ref{fig2:(a)}.

% \textcolor{blue}{To better understand the efficacy of INT4 block-wise PTQ than original FP checkpoint, Figure~\ref{fig2:weight_distribution} provides weight distribution and corresponding quantization bins before and after QAT training for both FP16 and INT4 PTQ cases, respectively. For FP16 case in (a), initial weight values are not spread enough to cover every quantization bins. It could prevent the model from utilizing its full representational capacity during training. Conversely, INT4 case in (b) provides evenly scattered weight distribution to cover every quantization bins. It enables to more flexible QAT, which also provides well-scatter weight distribution after training. Please see Appendix \ref{xx}, which provides further gradient-based analyses on weights and scales.}

One might question whether to leverage INT4 QAT instead of INT4 block-wise PTQ, considering that QAT typically outperforms PTQ. However, it is noteworthy that QAT requires several hundred million to billions of tokens and substantial computational resources$-$involving around one to two days with a single 8-GPU node for models in the 3B parameter range. In contrast, block-wise PTQ can be executed with just one to two million tokens sampled from a pre-training dataset such as C4~\citep{raffel2023exploring}, taking only a few hours with just a single GPU. Given this stark difference in resource requirements, block-wise PTQ is vastly more efficient$-$its computational cost is almost negligible compared to QAT$-$while still achieving comparable accuracy to the original FP16 baseline under INT4 per-channel quantization. As a result, we opt for INT4 block-wise PTQ over INT4 QAT.

\begin{table}[t]
\caption{Ablation results of OmniQuant and FlexRound, representative INT4 block-wise PTQ methods, on various benchmarks using Llama 3.2 3B Instruct after INT2 QAT with 5B training tokens. Scores for each task are reported as \textit{OmniQuant/FlexRound} (\textbf{Bold} means the best result).}
% \vspace{-5pt}
\label{tab:init-ablation}
\small
\begin{center}
\resizebox{0.999\linewidth}{!}{
\begin{tabular}{lccccc}
\toprule
Method & Bitwidth & WikiText2 $(\downarrow)$ & CSR Avg. $(\uparrow)$ & MMLU $(\uparrow)$ & IFEval $(\uparrow)$ \\
\midrule
INT4 PTQ & 4 & $12.52 / \mathbf{10.84}$ & $63.43/\mathbf{64.82}$ & $56.36/\mathbf{58.60}$ & $52.08/\mathbf{52.57}$ \\
\hdashline\noalign{\vskip 0.5ex}	
INT4 PTQ $\rightarrow$ NTP-QAT &  2 & $9.91/\mathbf{9.81}$ & $65.17/\mathbf{65.66}$ & $48.40/\mathbf{49.73}$ & $\mathbf{20.67}/20.51$ \\
\hdashline\noalign{\vskip 0.5ex}	
INT4 PTQ $\rightarrow$ Distill-QAT & 2 & $11.51/\mathbf{11.49}$ & $\mathbf{63.41}/63.04$ & $52.75/\mathbf{53.20}$ & $44.68/\mathbf{45.19}$ \\
\bottomrule
\end{tabular}
}
\end{center}
\vspace{-5pt}
\end{table}

\subsection{INT2 Distillation-based Quantization-Aware Training (Distill-QAT)}\label{subsec:distill_qat}

% Although INT2 QAT initialized from $\bm{W}_{\text{INT4}\rightarrow\text{INT2}}$ demonstrates strong performance on language understanding and reasoning tasks such as MMLU and commonsense reasoning benchmarks
Most existing QAT techniques~\citep{liu2023llmqat,chen2024efficientqat,liu2025paretoq} rely on next-token prediction (i.e., NTP-QAT). However, minimizing the next-token prediction loss on a pre-training corpus during INT2 NTP-QAT of instruction-tuned LLMs often presents challenges in recovering their instruction-following capability$-$a defining feature of instruction-tuned LLMs. This limitation stems from the fact that pre-training corpora primarily consist of general text rather than instruction-response pairs. To address this issue, we introduce INT2 Distill-QAT, which trains INT2 instruction-tuned LLMs to mimic the token-level probability distribution of their original FP16 counterparts, thereby allowing them to generate responses in a manner consistent with their FP16 versions.

To train INT2 instruction-tuned LLMs to imitate the token-level probability distribution of their FP16 baselines, INT2 Distill-QAT minimizes the generalized JSD between the INT2 quantized model (student, denoted as $\bm{W}_{\text{INT4}\rightarrow\text{INT2}}$) and its original FP16 counterpart (teacher, denoted as $\bm{W}_{\text{FP16}}$), which is a widely used divergence measure in LLM knowledge distillation~\citep{agarwal2024gkd,ko2024distillm}. 
More formally, let $P_{\bm{\Theta}}$ denote the conditional probability modeled by a decoder-only transformer parameterized by $\bm{\Theta}$. Given a pre-training token sequence $\mathcal{X} = \{x_1, \cdots, x_N\}$, the objective of INT2 Distill-QAT is given by
\begin{flalign}
&\mathcal{L}_{JSD(\beta)}={1 \over N}\sum\limits_{n=1}^N\mathcal{D}_{JSD(\beta)}(P_{\bm{W}_{\text{FP16}}}(\cdot|\mathcal{X}[\text{:}n])||P_{\bm{W}_{\text{INT4}\rightarrow\text{INT2}}}(\cdot|\mathcal{X}[\text{:}n])),\label{eq:distill_qat}&&\\
&\text{where }\mathcal{D}_{JSD(\beta)}(P_{\bm{W}_{\text{FP16}}}||P_{\bm{W}_{\text{INT4}\rightarrow\text{INT2}}}) = \beta\mathcal{D}_{KL}(P_{\bm{W}_{\text{FP16}}}||\beta P_{\bm{W}_{\text{FP16}}}+(1-\beta) P_{\bm{W}_{\text{INT4}\rightarrow\text{INT2}}})&&\nonumber\\
&\quad\quad\quad\quad\quad\quad\quad\quad\quad\quad\quad\quad\quad\quad\quad\,\,+(1-\beta)\mathcal{D}_{KL}(P_{\bm{W}_{\text{INT4}\rightarrow\text{INT2}}}||\beta P_{\bm{W}_{\text{FP16}}}+(1-\beta) P_{\bm{W}_{\text{INT4}\rightarrow\text{INT2}}}),&&\nonumber
\end{flalign}
$\mathcal{D}_{KL}$ is the KL-divergence, $\mathcal{X}[\text{:}n] = \{x_1, \cdots, x_{n-1}\}$, and $\beta$ is an interpolation coefficient between $0$ and $1$ (default: $0.5$). The rationale for selecting the generalized JSD is provided in Section~\ref{subsec:kd-loss-ablation}.

By minimizing the loss in Eq.~\ref{eq:distill_qat} with respect to $\bm{W}_{\text{INT4}}$ and $\bm{\Delta}_{\text{INT4}\rightarrow\text{INT2}}-$representing the model and quantization parameters of $\bm{W}_{\text{INT4}\rightarrow\text{INT2}}$, respectively$-$we ultimately quantize instruction-tuned LLMs to INT2 while preserving their instruction-following ability as evidenced in Table~\ref{tab:qual-ifeval}. We refer to this whole approach (i.e., INT4 PTQ $\rightarrow$ INT2 Distill-QAT) as UPQ. A notable aspect here is that during QAT$-$whether using NTP-QAT or Distill-QAT$-$$\bm{W}_{\text{INT4}}$ is treated as FP16 weights. In other words, although $\bm{W}_{\text{INT4}}$ is initially composed of $16$ discrete values, it is optimized as if it were in FP16, allowing it to evolve beyond the original $16$-value constraint over the course of QAT.

\begin{table}[t]
\caption{Ablation results of different distillation loss functions in the UPQ framework on various benchmarks using Llama 3.2 1B/3B Instruct models with 10B/5B training tokens (\textbf{Bold} indicates the best result, and \underline{underline} represents the second best result).}
\label{tab:loss-ablation}
\begin{center}
\small
% \tabcolsep=1em
% \resizebox{0.99\linewidth}{!}{
\begin{tabular}{lccccc}
\toprule
Method & WikiText2 $(\downarrow)$ & CSR Avg. $(\uparrow)$ & MMLU $(\uparrow)$ & IFEval $(\uparrow)$ \\
\midrule
Llama 3.2 1B Instruct (FP)                & $12.14$ &  $59.11$ & $45.46$ & $44.73$ \\[0.5ex]
\hdashline\noalign{\vskip 0.5ex}	
Confidence-aware KLD \citep{bitdistiller}  & $16.11$ & $56.31$ & $33.39$ & $27.44$ \\
Token-scaled KLD \citep{tokenscaled}  & $16.24$ & $54.64$ & $\underline{35.56}$ & $\underline{28.58}$ \\ 
Generalized JSD   & $\underline{15.97}$ & $\underline{56.47}$ & $\textbf{35.85}$ & $\textbf{30.51}$ \\
Generalized JSD + NTP & \textbf{$\textbf{14.78}$} & $\textbf{56.98}$ & $24.86$ & $20.84$ \\
\midrule
Llama 3.2 3B Instruct (FP)                & $10.48$ & $65.44$ & $59.92$ & $57.80$ \\[0.5ex]
\hdashline\noalign{\vskip 0.5ex}	
Confidence-aware KLD \citep{bitdistiller}  & $11.67$ & $\underline{63.70}$ & $53.19$ & $\underline{43.78}$ \\
Token-scaled KLD \citep{tokenscaled}  & $\underline{11.37}$ & $62.95$ & $\textbf{53.27}$ & $43.45$ \\ 
Generalized JSD   & $11.49$ & $63.04$ & $\underline{53.20}$ & $\textbf{45.19}$ \\
Generalized JSD + NTP & $\textbf{10.05}$ & $\textbf{66.68}$ & $50.76$ & $21.69$ \\
\bottomrule
\end{tabular}
% }
\end{center}
\vspace{-5pt}
\end{table}

\section{Experiments}\label{sec:exp}
In this section, we evaluate UPQ on a range of downstream task benchmarks.
As \citet{liu2025paretoq} demonstrates that NTP-QAT with SEQ (i.e., ParetoQ) substantially outperforms existing QAT techniques$-$such as BitDistiller~\citep{bitdistiller} and EfficientQAT~\citep{chen2024efficientqat}$-$at 2-bit precision on zero-shot CSR benchmarks, we primarily compare UPQ against NTP-QAT.
Furthermore, we focus on quantizing instruction-tuned LLMs$-$specifically, Llama 3.2 1B Instruct, Llama 3.2 3B Instruct, and Llama 3.1 8B Instruct~\citep{grattafiori2024llama3}$-$rather than pretrained LLMs. In addition, we are interested in preserving the capabilities of trained models, not in training models from scratch.

For Llama 3.2 1B Instruct, we follow a training schedule of 30B tokens, which corresponds to the saturation point reported by \citet{liu2025paretoq}.
Due to resource constraints, Llama 3.2 3B Instruct and Llama 3.1 8B Instruct are trained with shorter schedule of 5B tokens.
The pre-training dataset we used is DCLM-Edu~\citep{allal2025smollm2smolgoesbig}, which is filtered from DCLM~\citep{li2024datacomplm} by applying an educational quality classifier~\citep{lozhkov2024fineweb-edu}.
We further refined the dataset by retaining only those samples with a quality score greater than or equal to 3.
All training texts were packed with a context length of 1024 tokens.
Further details of our experimental settings are provided in Appendix. % ~\ref{subsec:additional-exp-details}.

To justify the efficacy of UPQ, we cover both easy and challenging downstream tasks.
The easy tasks include:  
(1) WikiText2 perplexity (PPL)~\citep{merity2016pointer}, and  
(2) Average on five zero-shot CSR benchmarks (CSR Avg.): ARC-e, ARC-c~\citep{allenai:arc}, PIQA~\citep{Bisk2020}, HellaSwag~\citep{zellers2019hellaswag}, and WinoGrande~\citep{ai2:winogrande}.  
The challenging tasks include:  
(1) MMLU~\citep{hendrycks2021mmlu}, and  
(2) IFEval~\citep{zhou2023ifeval}, both of which, to our knowledge, are being simultaneously investigated for the first time in LLM quantization without training from scratch.% —neither of which, to our knowledge, have been addressed in the context of 2-bit PCQ in prior literature.
WikiText2 PPL is evaluated at a 4096 context length to confirm no degradation on longer sequences unseen during QAT.
The other benchmarks are conducted using the Language Model Evaluation Harness~\citep{eval-harness}, and we retain its default settings.

Our results demonstrate that UPQ achieves state-of-the-art performance on INT2 quantization of instruction-tuned LLMs.
Furthermore, qualitative analysis suggests that the instruction-following ability significantly deteriorates without our method.
To validate the effectiveness of UPQ, we also conduct ablation studies on:  
(1) INT4 PTQ methods, and  
(2) distillation loss functions.

% \subsection{Initialization PTQ Method ablation Study}\label{subsec:init-ptq-method-ablation}
% \input{tables/4_init_ablation}
% \input{figures/7_ptq_ablation}

\subsection{Ablation Study}\label{subsec:kd-loss-ablation}
\paragraph{INT4 PTQ Method Study} We compare FlexRound and OmniQuant, as described in Section~\ref{subsec:blockwise}, after INT2 QAT (both NTP-QAT and Distill-QAT). Table~\ref{tab:init-ablation} shows that FlexRound slightly outperforms OmniQuant on most benchmarks across PTQ, NTP-QAT, and Distill-QAT. Based on this observation, we adopt FlexRound as the default method for INT4 block-wise PTQ, unless otherwise specified.

\paragraph{Distillation Loss Study} We conduct an ablation study of various distillation loss functions in UPQ. Generalized JSD in Eq.~\ref{eq:distill_qat} is compared with Confidence-Aware KL Divergence loss from BitDistiller and Token-Scaled Logit Distillation loss. Additionally, we include Generalized JSD + NTP, to evaluate the effect of mixing two different losses. Table \ref{tab:loss-ablation} indicates that Generalized JSD consistently improves performance on MMLU and IFEval compared to other loss functions. Generalized JSD + NTP surpasses Generalized JSD on WikiText2 and CSR Avg., but shows degraded performance on MMLU and IFEval. Hence, we choose Generalized JSD as the default loss function in Distill-QAT.

\subsection{Main Results}\label{subsec:main-results}

\begin{table}[t]
\caption{Benchmark results of four INT2 QAT methods applied to Llama 3.2 1B Instruct, Llama 3.2 3B Instruct, Llama 3.1 8B Instruct.}
\label{tab:all-tasks}
\begin{center}
\small
% \tabcolsep=1em
% \resizebox{0.99\linewidth}{!}{
\begin{tabular}{lccccc}
\toprule
Method & \# tokens & WikiText2 $(\downarrow)$ & CSR Avg. $(\uparrow)$ & MMLU $(\uparrow)$ & IFEval $(\uparrow)$ \\
\midrule
Llama 3.2 1B Instruct                 & NA & $12.14$ & $59.11$ & $45.46$ & $44.73$ \\
\hdashline	
NTP-QAT                                  & 30B & $14.86$ & $\mathbf{59.81}$ & $27.03$ & $20.87$ \\ % ParetoQ
Distill-QAT               & 30B & $18.35$ & $55.54$ & $33.33$ & $27.84$ \\ % No PTQ init. + JSD
INT4 PTQ $\rightarrow$ NTP-QAT            & 30B & $\mathbf{14.46}$ & $59.25$ & $25.37$ & $20.50$ \\ % ParetoQ
UPQ (Ours)                           & 30B & $15.46$ & $56.18$ & $\mathbf{37.59}$ & $\mathbf{28.56}$ \\
\midrule
Llama 3.2 3B Instruct                 & NA & $10.48$ & $65.44$ & $59.92$ & $57.80$ \\
\hdashline
NTP-QAT                                  & 5B & $11.96$ & $60.94$ & $39.17$ & $19.97$ \\ % ParetoQ 
Distill-QAT               & 5B & $16.18$ & $59.01$ & $45.29$ & $27.12$ \\
INT4 PTQ $\rightarrow$ NTP-QAT & 5B   & $\mathbf{9.81}$ & $\mathbf{65.66}$ & $49.73$ & $20.97$ \\ % ParetoQ
UPQ (Ours)                           & 5B & $11.49$ & $63.04$ & $\mathbf{53.20}$ & $\mathbf{45.19}$ \\
\midrule
Llama 3.1 8B Instruct & NA & $6.75$ & $73.72$ & $68.21$ & $50.05$ \\
\hdashline
NTP-QAT & 5B & $14.31$ & $64.42$ & $43.35$ & $20.81$\\
Distill-QAT & 5B & $10.69$ & $67.82$ & $54.39$ & $30.99$ \\
INT4 PTQ $\rightarrow$ NTP-QAT & 5B & $\mathbf{8.36}$ & $70.80$ & $55.81$ & $20.06$ \\
UPQ (Ours) & 5B & $8.42$ & $\mathbf{71.61}$ & $\mathbf{61.73}$ & $\mathbf{44.48}$ \\
\bottomrule
\end{tabular}
% }
\end{center}
\vspace{-5pt}
\end{table}

\begin{figure}[t]
    \centering
    \includegraphics[width=0.99\linewidth]{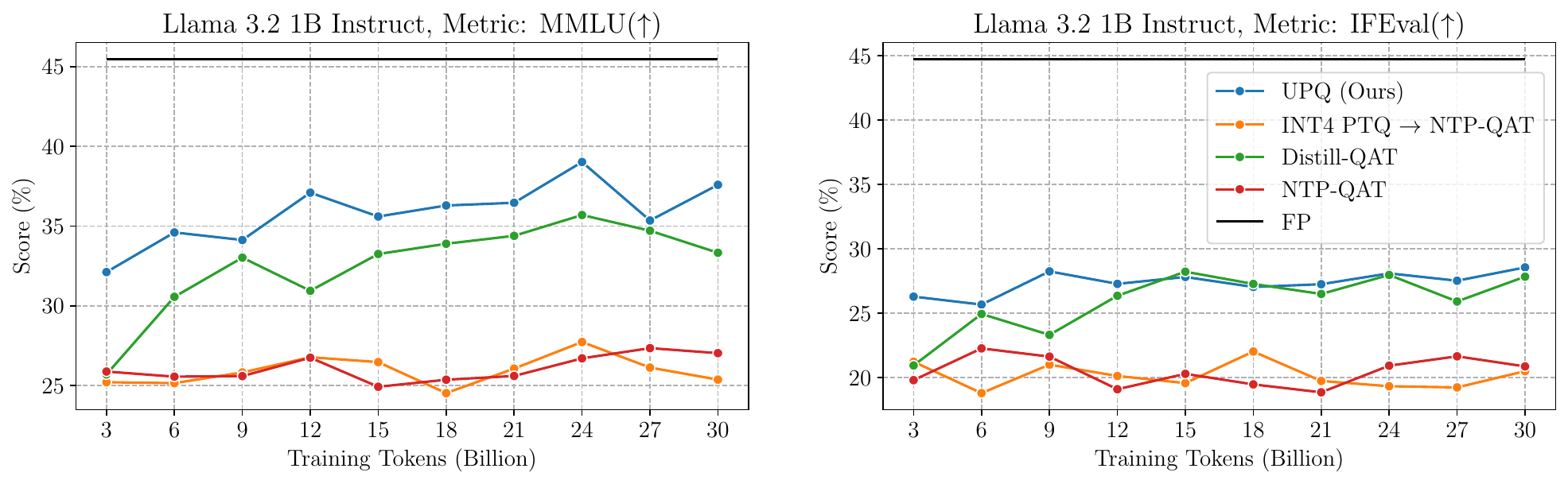}\label{fig:llama-1b-full-training}
    \caption{Change in MMLU (left) and IFEval (right) scores during training (up to 30B tokens) depending on four INT2 QAT methods. All metrics were obtained with Llama 3.2 1B Instruct.}
    \vspace{-5pt}
\end{figure}

In our main results, we compare four QAT methods: (1) \textbf{NTP-QAT}, (2) \textbf{Distill-QAT}, (3) \textbf{INT4 PTQ $\rightarrow$ NTP-QAT}, and (4) \textbf{UPQ} (ours).
This experimental setup is designed to demonstrate that both techniques proposed in Sections~\ref{subsec:blockwise} and~\ref{subsec:distill_qat} should be integrated to effectively recover the intrinsic capabilities of instruction-tuned LLMs.

Let us begin with Figure~\ref{fig:llama-1b-full-training}. According to \citet{liu2025paretoq}, the CSR average score saturates at 30B training tokens under NTP-QAT. However, to our surprise, we observe that neither NTP-QAT nor INT4 PTQ $\rightarrow$ NTP-QAT yields any improvement on Llama 3.2 1B Instruct in MMLU or IFEval scores. For instance, MMLU accuracy remains around 25\%, akin to random guessing.
These results suggest that NTP alone is insufficient to restore general language understanding and instruction-following after severe quantization (e.g. 2-bit per-channel). The core abilities of instruction-tuned LLMS remains unrepaired even with extensive training up to 30B tokens.

Table~\ref{tab:all-tasks} broadens this observation by comparing the four QAT methods across Llama 3.2 1B Instruct, Llama 3.2 3B Instruct, and Llama 3.1 8B Instruct.
Across all model sizes, UPQ consistently outperforms the others on the MMLU and IFEval benchmarks.
Notably, IFEval scores completely collapsed under both NTP-QAT and INT4 PTQ $\rightarrow$ NTP-QAT.
This underscores that distillation is a key component for QAT of instruction-tuned LLMs.

In contrast, our strategy$-$starting from INT4 block-wise PTQ$-$yields substantial improvements in MMLU and IFEval scores over the naive initialization.
This improvement stand out especially in the larger models (3B or 8B).
For instance, in Llama 3.2 3B Instruct, the MMLU score and the IFEval score improve from $45.29$ to $53.20$ and from $27.12$ to $45.29$ respectively.
Similary, in Llama 3.1 8B Instruct, the MMLU score increases from $54.39$ to $61.73$, and the IFEval score improves from $30.99$ to $44.48$.
Even on easy downstream tasks such as WikiText2 and CSR Avg., INT4 PTQ $\rightarrow$ NTP-QAT-combining our initialization strategy with NTP-proves effective, with only one exception: the CSR Avg. score of Llama 3.2 1B Instruct under NTP-QAT.
This demonstrates that a well-chosen initialization could recover the degradation of instruction-following behavior, even without relying on post-training-style datasets typically employed in building instruct-tuned LLMs. 

The details of instruction-following behavior across the QAT methods are shown in Table~\ref{tab:qual-ifeval}, which presents qualitative results for Llama 3.2 3B Instruct on the IFEval benchmark. 
While we examined many qualitative examples (see Appendix), consistent patterns emerge across model behaviors: 1) NTP-QAT and INT4 PTQ $\rightarrow$ NTP-QAT tend to produce repetitive outputs early in the generation process, and 2) Distill-QAT is more likely to follow the instruction initially but tends to fall into repetition midway through the generation process more often than UPQ. % ~\ref{subsec:additional-ifeval}

\subsection{Analysis of Learnable Parameter Dynamics during Distill-QAT and UPQ} \label{subsec:analysis-param-dynamics}
\begin{figure}
    \begin{center}
        \includegraphics[width=\textwidth]{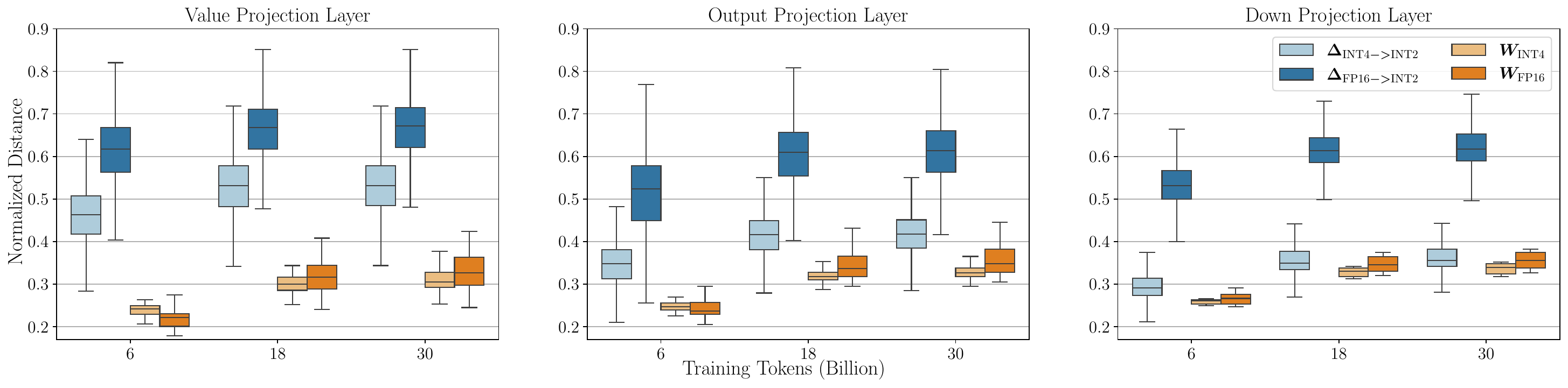}
    \end{center}
    \vspace{-5pt}
    \caption{Normalized L1 distance dynamics of learnable parameters $\bm{\Delta}_{\textrm{FP16} \rightarrow \textrm{INT2}}$ and $\bm{W}_{\textrm{FP16}}$ (in Eq.~\ref{eq:original_seq}) during Distill-QAT, and $\bm{\Delta}_{\textrm{INT4} \rightarrow \textrm{INT2}}$ and $\bm{W}_{\textrm{INT4}}$ (in Eq.~\ref{eq:modified_seq}) during UPQ of Llama 3.2 1B Instruct (Value, Output, and Down projection layers). The statistics are aggregated across all layers, respectively. Note that both $\bm{W}_{\textrm{INT4}}$ and $\bm{W}_{\textrm{FP16}}$ are normalized by the original model weights.}
    \label{fig:scale-weight-dynamics-1}
    \vspace{-5pt}
\end{figure}

%%% Text for Figure \ref{fig:scale-weight-dynamics-1} %%%
It is known that low-bit QAT (e.g., $\leq$ 2-bit) tends to exhibit a "\textit{reconstruction}" behavior rather than "\textit{compensation}"~\citep{liu2025paretoq}.
However, when the goal is to recover the intrinsic behavior of instruction-tuned LLMs from quantization error, training dynamics resembling "\textit{reconstruction}" should be avoided particularly.
This is because instruction-tuned LLMs are meticulously fine-tuned, making them more susceptible to behavioral degradation when their parameters are altered drastically.

From this perspective, Figure~\ref{fig:scale-weight-dynamics-1} illustrates dynamics of learnable parameters during QAT.
It shows that our initialization strategy encourages weight updates that are more "\textit{compensatory}" in nature. $\bm{\Delta}_{\textrm{INT4} \rightarrow \textrm{INT2}}$ consistently deviates less than $\bm{\Delta}_{\textrm{FP16} \rightarrow \textrm{INT2}}$ during training.
Although $\bm{W}_{\textrm{INT4}}$ starts with greater deviation than $\bm{W}_{\textrm{FP16}}$ due to the initial PTQ, both converge to a similar level as training progresses.
This phenomenon offers insight into why our initialization strategy is effective for QAT of instruction-tuned LLMs.

\section{Conclusion}

We propose UPQ, a progressive quantization framework that first quantizes an FP16 instruction-tuned LLM to INT4 using block-wise PTQ, and then to INT2 using Distill-QAT. Our proposed method utilizes only public data, without relying on proprietary sources, to successfully quantize most popular open-source instruction-tuned LLMs ranging from 1B to 8B parameters. The resulting INT2 quantized models recover strong language understanding, reasoning, and instruction-following performance, as shown on the MMLU and IFEval benchmarks.
%Without relying on proprietary post-training data, UPQ successfully quantizes open-source instruction-tuned LLMs ranging from 1B to 8B parameters to INT2, while recovering language understanding, reasoning, and instruction-following performance, as demonstrated on the MMLU and IFEval benchmarks.We hope this work paves the way for the practical adoption of INT2 quantization in instruction-tuned LLMs.

\newpage
% \section*{References}

{
\small
\bibliographystyle{unsrtnat}
\bibliography{references}

% [1] Alexander, J.A.\ \& Mozer, M.C.\ (1995) Template-based algorithms for
% connectionist rule extraction. In G.\ Tesauro, D.S.\ Touretzky and T.K.\ Leen
% (eds.), {\it Advances in Neural Information Processing Systems 7},
% pp.\ 609--616. Cambridge, MA: MIT Press.

% [2] Bower, J.M.\ \& Beeman, D.\ (1995) {\it The Book of GENESIS: Exploring
%   Realistic Neural Models with the GEneral NEural SImulation System.}  New York:
% TELOS/Springer--Verlag.

% [3] Hasselmo, M.E., Schnell, E.\ \& Barkai, E.\ (1995) Dynamics of learning and
% recall at excitatory recurrent synapses and cholinergic modulation in rat
% hippocampal region CA3. {\it Journal of Neuroscience} {\bf 15}(7):5249-5262.
}

%%%%%%%%%%%%%%%%%%%%%%%%%%%%%%%%%%%%%%%%%%%%%%%%%%%%%%%%%%%%

\newpage
\appendix
% \section{Technical Appendices and Supplementary Material}

\section{Next-Token Prediction-based Qantization-Aware Training (NTP-QAT)}\label{appendix:ntp-qat}

Let $P_{\bm{\Theta}}$ denote the conditional probability modeled by a decoder-only transformer parameterized by $\bm{\Theta}$. Given a pre-training token sequence $\mathcal{X} = \{x_1, \cdots, x_N\}$, the objective of INT2 NTP-QAT is given by
\begin{flalign}
\mathcal{L}_{NTP}={1 \over N}\sum\limits_{n=1}^N\log P_{\bm{W}_{\text{FP16}\rightarrow\text{INT2}}}(x_n|x_1,\cdots,x_{n-1}),\label{eq:NTP-QAT} % \text{ without using INT4 block-wise PTQ}
\end{flalign}
or
\begin{flalign}
\mathcal{L}_{NTP}={1 \over N}\sum\limits_{n=1}^N\log P_{\bm{W}_{\text{INT4}\rightarrow\text{INT2}}}(x_n|x_1,\cdots,x_{n-1}) \label{eq:INT4-PTQ-NTP-QAT}, % \text{ with using INT4 block-wise PTQ}.
\end{flalign}
depending on whether INT4 block-wise PTQ is employed or not. When minimizing the loss in Eq.~\ref{eq:NTP-QAT} with respect to $\bm{W}_{\text{FP16}}$ and $\bm{\Delta}_{\text{FP16}\rightarrow\text{INT2}}-$representing the model and quantization parameters of $\bm{W}_{\text{FP16}\rightarrow\text{INT2}}$, respectively$-$we refer to this approach as NTP-QAT, which is identical ParetoQ~\citep{liu2025paretoq}. In a similar manner to Section~\ref{subsec:distill_qat}, minimizing the loss in Eq.~\ref{eq:INT4-PTQ-NTP-QAT} with respect to $\bm{W}_{\text{INT4}}$ and $\bm{\Delta}_{\text{INT4}\rightarrow\text{INT2}}$ is termed \text{INT4 PTQ $\rightarrow$ NTP-QAT}.

\newpage
\section{Weight Distribution in Llama 3.2 3B Instruct before and after NTP-QAT} \label{subsec:additional-weight-distribution}
\begin{figure}[h]
    \subfigure[Weight distribution before (above) and after (below) INT2 NTP-QAT, starting from original FP16 weights, $\bm{W}_{\text{FP16}}$]{\includegraphics[width=0.348\linewidth]{
    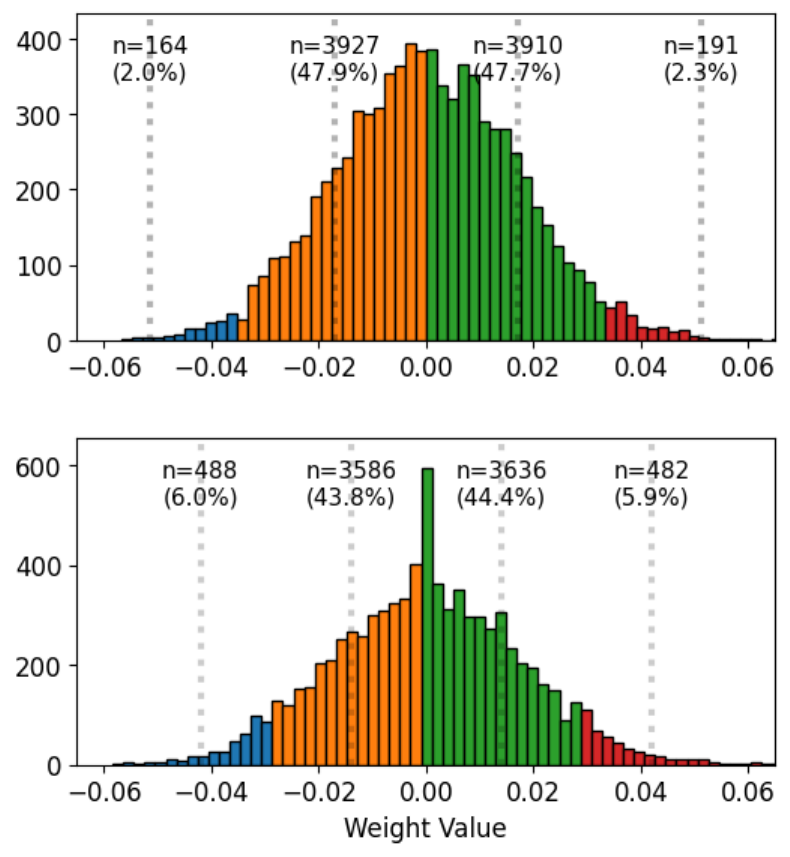}\label{fig5:(a)}
    }
    \hspace{0.0025\textwidth}
    \subfigure[Weight distribution before (above) and after (below) INT2 NTP-QAT, starting from INT4 PTQ weights, $\bm{W}_{\text{INT4}}$]{\includegraphics[width=0.348\linewidth]{
    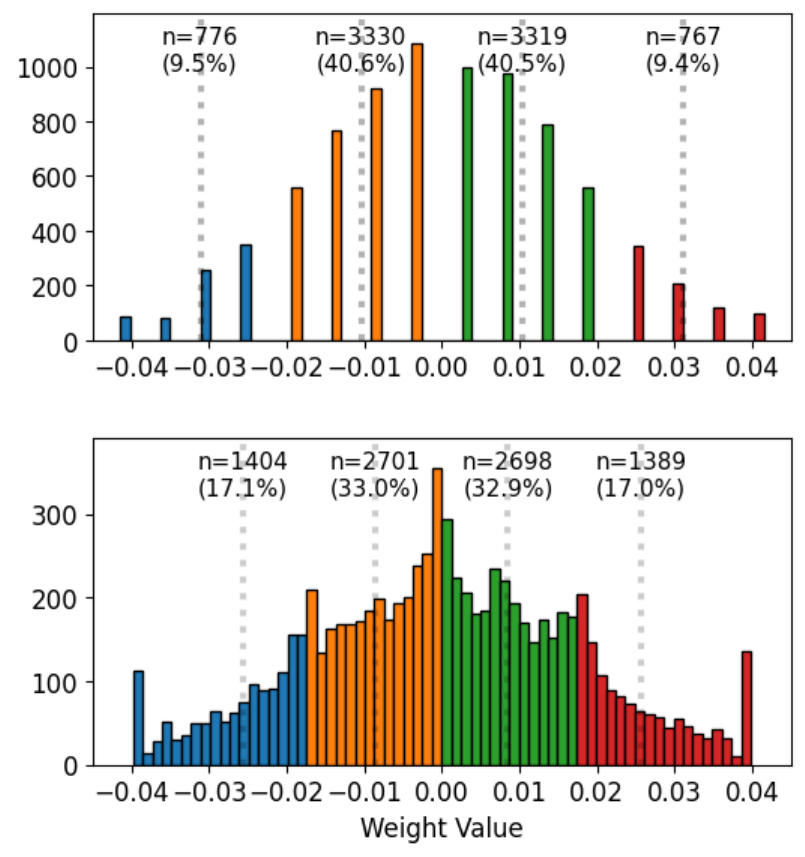}\label{fig5:(b)}
    }
    \hspace{0.0025\textwidth}
    \subfigure[Training loss curves of INT2 NTP-QAT and INT4 PTQ $\rightarrow$ NTP-QAT]{\includegraphics[width=0.26\linewidth]{
    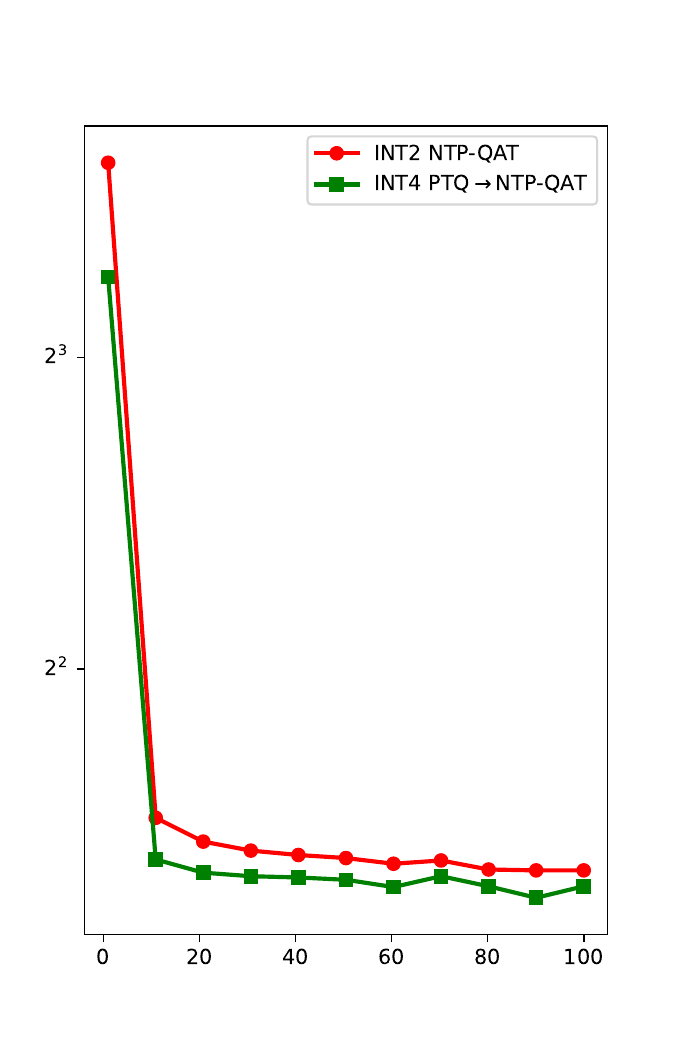}\label{fig5:(c)}
    % illustrates the dynamics of additional learnable parameters during QAT, specifically those in the Query, Key, Up, and Gate projection layers, which are not covered in Figure 4.
    }
    % \vspace{-5pt}
    \caption{Weights distribution within the first channel of the first down-projection layer in Llama 3.2 3B Instruct. Dotted lines denote four quantization levels of 2-bit, and the corresponding weights are differently colored.}
    \label{fig5:weight_distribution_ntp}
\end{figure}
Similar to Figure~\ref{fig2:weight_distribution}, we investigate the weight distribution change in Figure~\ref{fig5:weight_distribution_ntp} when utilizing the next-token prediction loss in Section~\ref{appendix:ntp-qat} instead of the generalized JSD loss. As with the case with the generalized JSD loss in Figure~\ref{fig2:weight_distribution}, initializing from INT4 PTQ weights results in increased utilization of the large-magnitude INT2 quantization bins (i.e., $\{-3,3\}$) than initializing from FP16 weights. This demonstrates that starting QAT from INT4 PTQ weights is beneficial across different loss functions. 

\newpage
\section{Additional qualitative evaluation on IFEval}\label{subsec:additional-ifeval}
% \usepackage{multirow}
% \usepackage{graphicx}
\begin{table}[ht]
\caption{Additional qualitative evaluation of Llama 3.2 3B Instruct on IFEval after four INT2 QAT techniques with 5B tokens. \textcolor{orange}{Orange} highlights repetitive generation upon reaching the maximum token limit; \textcolor{blue}{blue} and \textcolor{red}{red} indicate correct and incorrect instruction following, respectively.}
\resizebox{\textwidth}{!}{%
\begin{tabular}{m{0.3\textwidth} c m{0.765\textwidth}}
\hline
\multicolumn{1}{c}{\textbf{Prompt}} &  \textbf{Method}           & \multicolumn{1}{c}{\textbf{Response}}                                                                                                                                                                                                                                                                                                                       \\ \hline
\multirow[c]{4}{0.3\textwidth}{\parbox{0.3\textwidth}{Write me a letter in the style of Shakespeare about the mandates and instructions of the King. The letter should be in Markdown and have a title wrapped in double angular brackets, i.e. <<title>>.\\\\Instruction 1: Markdown\\Instruction 2: <<title>>}} & \makecell{NTP-QAT\\(IFEval: 19.97)}                       & \text{ } The title should be in bold and the letter should be in italics. The letter should be 1-2 pages long and should be written in a formal style. \textcolor{orange}{The letter should be written in a formal style and should not be written in a casual style. The letter should be written in a formal style and should not be written in a casual style.$\cdots$\textit{(repeated)}} $\Rightarrow$  (\textcolor{red}{Instruction 1}, \textcolor{red}{Instruction 2})                                                                                                                                                                                                           \\ \cline{2-3} 
                                                                                                         & \makecell{Distill-QAT\\(IFEval: 27.12)}    & \text{ }\textbackslash{}n\textbackslash{}nHere is the letter:\textbackslash{}n\textbackslash{}n```\textbackslash{}n\text{\#\#} Letter from King Henry to his subjects\textbackslash{}n\textbackslash{}nDear my dear subjects,\textbackslash{}n\textbackslash{}nI, King Henry, have been entrusted with the great responsibility of governing our kingdom. As such, I have been tasked with the duty of ensuring the well-being and prosperity of my people.\textbackslash{}n\textbackslash{}nIn this letter, I shall outline the mandates and instructions that I have been given by the King of England. I shall also provide guidance on how to fulfill these tasks and responsibilities.\textbackslash{}n\textbackslash{}n**Mandates and Instructions**\textbackslash{}n\textbackslash{}n1.  **The Royal Writ**: The Royal Writ is a sacred document that holds the power to grant or deny royal favors. It is a powerful tool that should be used with great care and caution.\textbackslash{}n2.  **The Royal Seal**: The Royal Seal is a symbol of the King's authority and power. It should be used sparingly and only in situations where it is absolutely necessary.\textbackslash{}n3.  **The Royal Writs**: The Royal Writs are a series of documents that are issued by the King to grant or deny royal favors. They should be used sparingly and only in situations where it is absolutely necessary. $\cdots$ $\Rightarrow$ (\textcolor{blue}{Instruction 1}, \textcolor{red}{Instruction 2})                                                     \\ \cline{2-3} 
                                                                                                         & \makecell{INT4 PTQ $\rightarrow$ NTP-QAT\\(IFEval: 20.97)} & \text{ }The letter should be at least 100 words long and should be written in the style of Shakespeare. The letter should be written in the first person and should be addressed to the King. \textcolor{orange}{The letter should be written in the style of Shakespeare and should be written in the first person. The letter should be written in the style of Shakespeare and should be written in the first person.$\cdots$\textit{(repeated)}} $\Rightarrow$ (\textcolor{red}{Instruction 1}, \textcolor{red}{Instruction 2}) \\ \cline{2-3} 
                                                                                                         & \makecell{UPQ (Ours)\\(IFEval: 45.19)}                & 
\text{ }<<title>>: A Royal Decree\textbackslash{}n\textbackslash{}nTo my most trusted and loyal subjects,n\textbackslash{}nI, the King, do hereby decree that all citizens of my realm shall adhere to the following mandates and instructions:\textbackslash{}n\textbackslash{}n*   **The Royal Decree**: All citizens shall attend the weekly town hall meetings, where they shall be informed of the latest news and decrees from the throne.\textbackslash{}n*   **The Royal Decree**: All citizens shall maintain a clean and tidy abode, free from filth and disorder.\textbackslash{}n*   **The Royal Decree**: All citizens shall respect the authority of the nobles and the clergy, and shall not question their decisions or actions.\textbackslash{}n*   **The Royal Decree**: All citizens shall provide for the well-being and safety of their families, and shall not neglect their duties to the state.\textbackslash{}n*   **The Royal Decree**: All citizens shall be mindful of the environment and shall take steps to protect and preserve the natural resources of the realm.\textbackslash{}n\textbackslash{}nThese mandates and instructions are not optional, but are a requirement for all citizens of the realm. Failure to comply shall result in severe penalties, including but not limited to fines, imprisonment, and even death.\textbackslash{}n\textbackslash{}nI, the King, command you to obey these decrees and to uphold the authority of the throne. May the blessings of the gods be upon you, and may your lives be filled with prosperity and happiness.\textbackslash{}n\textbackslash{}n---\textbackslash{}n\textbackslash{}nYours truly,\textbackslash{}nThe King\textbackslash{}n\textbackslash{}n(Note: I've tried to emulate the style of Shakespeare's letters, but I'm not a professional, and the language may not be entirely accurate. I've also tried to make the letter sound formal and authoritative, as would be expected from a monarch.) $\Rightarrow$ (\textcolor{blue}{Instruction 1}, \textcolor{blue}{Instruction 2})                          \\ \hline
\end{tabular}%
}
\label{tab:appendix-qual-ifeval}
\vspace{-5pt}
\end{table}
We provide additional qualitative evaluation on IFEval, as detailed in Table~\ref{tab:appendix-qual-ifeval}, to substantiate that UPQ can produce responses of higher quality than other QAT techniques. Similar to the observation in Table~\ref{tab:qual-ifeval}, only UPQ demonstrates consistent adherence to prompt instructions, thus attaining the highest score on IFEval.

\newpage
\section{Additional Figure of Normalized L1 Distance Dynamics}\label{subsec:additional-dynamics}
\begin{figure}[h]
    \begin{center}
        \includegraphics[width=\textwidth]{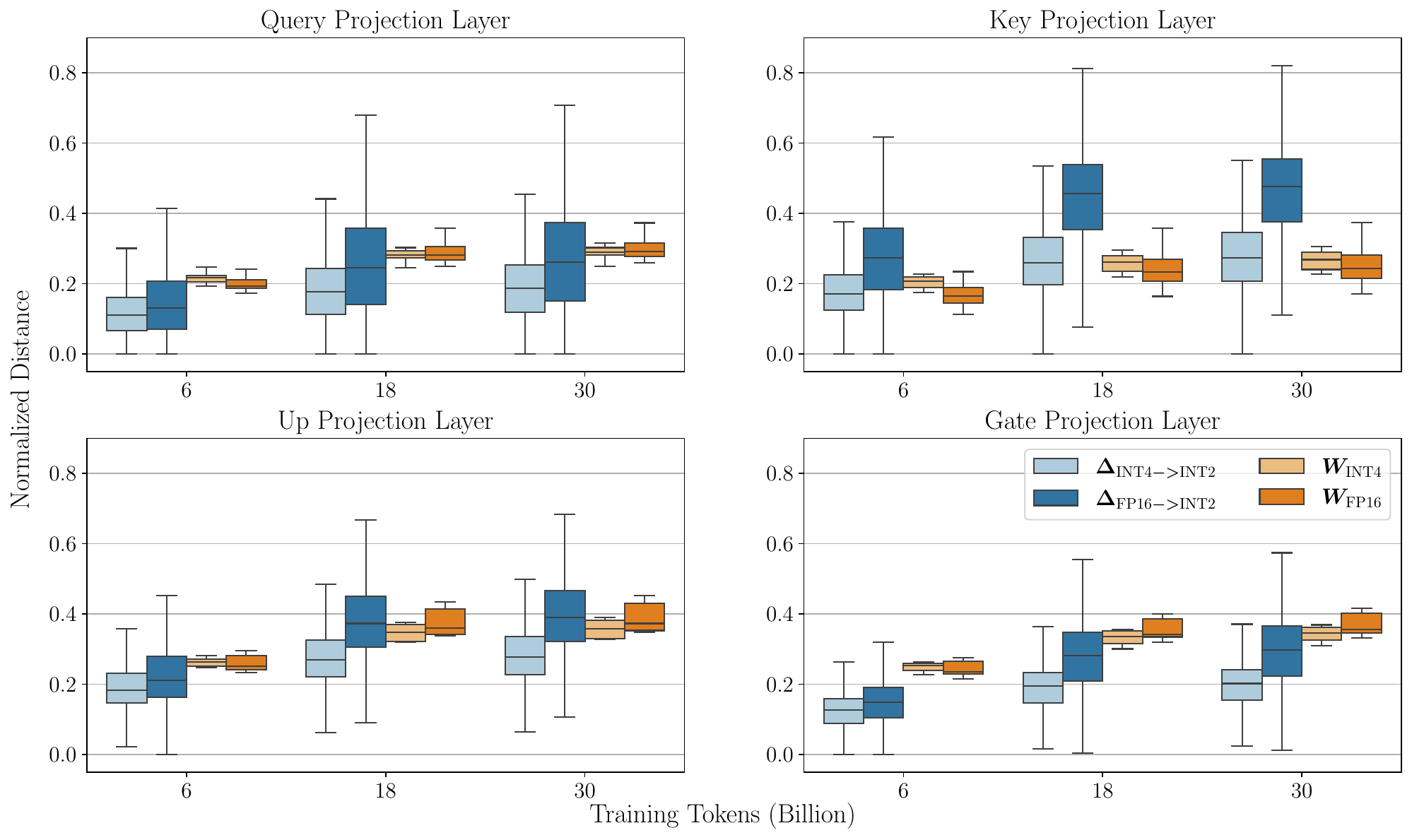}
    \end{center}
    \caption{Normalized L1 distance dynamics of learnable parameters $\bm{\Delta}_{\textrm{FP16} \rightarrow \textrm{INT2}}$ and $\bm{W}_{\textrm{FP16}}$ (in Eq.~\ref{eq:original_seq}) during Distill-QAT, and $\bm{\Delta}_{\textrm{INT4} \rightarrow \textrm{INT2}}$ and $\bm{W}_{\textrm{INT4}}$ (in Eq.~\ref{eq:modified_seq}) during UPQ of Llama 3.2 1B Instruct (Query, Key, Up and Gate projection layers). The statistics are aggregated across all layers, respectively. Note that both $\bm{W}_{\textrm{INT4}}$ and $\bm{W}_{\textrm{FP16}}$ are normalized by the original model weights.}
    \label{fig:scale-weight-dynamics-2}
\end{figure}

Figure~\ref{fig:scale-weight-dynamics-2} illustrates the dynamics of learnable parameters during QAT, specifically those in the Query, Key, Up, and Gate projection layers, which are not covered in Figure~\ref{fig:scale-weight-dynamics-1}.
Like in Figure~\ref{fig:scale-weight-dynamics-1}, $\bm{\Delta}_{\textrm{INT4} \rightarrow \textrm{INT2}}$ exhibits smaller changes, on average, in normalized L1 distance compared to $\bm{\Delta}_{\textrm{FP16} \rightarrow \textrm{INT2}}$.
Meanwhile, both $\bm{W}_{\textrm{INT4}}$ and $\bm{W}_{\textrm{FP16}}$ converge to similar levels by the end of training.
This behavior corresponds to the "\textit{compensatory}" dynamics previously discussed in Section~\ref{subsec:analysis-param-dynamics}.

\newpage
\section{Further Details of Our Experimental Settings} \label{subsec:additional-exp-details}

% Table
All experiments are performed on a single compute node equipped with 8 NVIDIA A100 GPUs.
We use the AdamW optimizer with zero weight decay, a learning rate of $2 \times 10^{-5}$ with cosine scheduling, and a total batch size of 256 per optimizer step.
Gradient accumulation is employed when GPU memory constraints prevent using the full batch size of 256 directly.
For Distill-QAT and UPQ, we use $\beta=0.5$ in Eq.~\ref{eq:distill_qat}.

\newpage
\section{Review on Further Quantization Methods} \label{subsec:review}
In this section, we briefly summarize notable quantization methods, which are not referred in Section~\ref{sec:related}. \textbf{AdaRound} \citep{nagel2020adaround} suggests an adaptive rounding method for PTQ, which optimizes weight quantizer by deciding whether each weight should be rounded up or down, instead of rounding-to-nearest. \textbf{BRECQ} \citep{li2021brecq} suggests a PTQ framework that performs block-wise reconstruction using second-order error analysis, and it balances cross-layer dependencies with per-layer sensitivity. For further efficient PTQ procedure, \textbf{GPTQ} \citep{frantar2022gptq} suggests a one-shot PTQ method which utilizes approximated second-order information to minimize the quantization error.

As a different direction, mixed-precision quantization methods \citep{wang2019haq, pandey2023practical} have been suggested to enable more flexible quantization by accounting for the sensitivity of parameters to quantization error. \textbf{AWQ} \citep{lin2023awq} identifies and rescales the most important weight channels based on activation sensitivity, thereby protecting salient weights to FP16 and enabling accurate 4-bit quantization without any fine-tuning or backpropagation. \textbf{SpQR} \citep{dettmers2023spqr} identifies few outlier weight by utilizing defined parameter sensitivity value, and it also stores them in higher precision while quantizing the rest. \textbf{GWQ} \citep{shao2024gwq} leverages gradient-based sensitivity analysis on a small calibration set to identify most important weights. 

Several studies have been proposed to effectively quantize not only weights but also activations, aiming to achieve end-to-end low-bit inference without performance degradation. \textbf{SmoothQuant} \citep{xiao2023smoothquant} mitigates activation outliers by transforming them into the weight domain via an equivalent transformation, enabling 8-bit activation quantization with negligible accuracy drop. \textbf{QDrop} \citep{Wei:ICLR22} utilizes dropout-like method, which drops activation quantization during calibration, encouraging a flatter loss landscape and improving robustness for low-bit quantization. \textbf{QuaRot} \citep{ashkboos2024quarot} introduces a new quantization scheme based on rotations, which removes outliers from the hidden state without changing the output, making quantization easier. As a variant of rotation-based method, \textbf{SpinQuant} \citep{Liu:ICLR25} introduces a training of rotation matrices into the PTQ process, preconditioning weight and activation distributions to remove outliers. \textbf{FlatQuant} \citep{Sun:ICML25} applies learnable affine transformations to each layer’s weights and activations, flattening their distributions to mitigate the impact of outliers.

\newpage
% \section{Recap : SEQ-based INT2 Quantization}
% Let \(\bm{W}_{\text{FP16}}\in \mathbb{R}^{m\times n}\) denote the FP16 weights,  
% and let \(\bm{\Delta}_{\text{FP16}\rightarrow\text{INT2}}\in\mathbb{R}^{m\times 1}\) be a per-channel learnable scale (initialized to the max absolute value across each row or channel).  
% Following the ParetoQ \citep{liu2025paretoq}, the INT2-quantized weights are obtained as:
% \begin{align}
% \label{eq:seq_def}
% \bm{W}_{\text{FP16}\rightarrow\text{INT2}}
% ~=~
% \frac{\bm{\Delta}_{\text{FP16}\rightarrow\text{INT2}}}{2}\,
% \Bigl(\,
% \Bigl\lfloor 
% 2~\text{clip}\Bigl(
% \tfrac{\bm{W}_{\text{FP16}}}{\bm{\Delta}_{\text{FP16}\rightarrow\text{INT2}}}
% ,\,-1+\epsilon,\,
% 1-\epsilon
% \Bigr)
% ~-~ 0.5 
% \Bigr\rceil
% ~+~
% 0.5
% \Bigr).
% \end{align}
% Here, 
% \(\text{clip}(x,a,b)=\min\{\max\{x,a\},\,b\}\), 
% and \(\epsilon>0\) is a small constant (e.g.\ 0.01).  
% As a result, each weight in \(\bm{W}_{\text{FP16}\rightarrow\text{INT2}}\) is mapped to one of \(\{\,-1.5\,\bm{\Delta},\,-0.5\,\bm{\Delta},\,0.5\,\bm{\Delta},\,1.5\,\bm{\Delta}\}\),  
% where \(\bm{\Delta}\equiv \bm{\Delta}_{\text{FP16}\rightarrow\text{INT2}}\).
\section{Gradient Analysis on Weight and Scale}
% Let \(\bm{W}\equiv \bm{W}_{\text{FP16}}\) and \(\bm{\Delta}\equiv \bm{\Delta}_{\text{FP16}\rightarrow\text{INT2}}\) for shorthand.
In this section, we denote $\bm{W}_{\text{FP16}}$ and $\bm{\Delta}_{\text{FP16}\rightarrow\text{INT2}}$ in Eq.~\ref{eq:original_seq} as $\bm{W}$ and $\bm{\Delta}$ for shorthand.
\subsection{Gradient with respect to Weight}
Define
\[
z 
~\coloneqq~
\text{clip}\!\Bigl(\tfrac{\bm{W}}{\bm{\Delta}},\, -1+\epsilon,\,1-\epsilon\Bigr), 
\quad
x
~=~
2\,z ~-~0.5.
\]
Then from \eqref{eq:original_seq}, 
\(\bm{W}_{\text{FP16}\rightarrow\text{INT2}} = \tfrac{\bm{\Delta}}{2}\,\Bigl(\lfloor x\rceil +0.5\Bigr)\).
\paragraph{Chain rule decomposition.}
We wish to compute
\[
\frac{\partial \,\bm{W}_{\text{FP16}\rightarrow\text{INT2}}}{\partial \,\bm{W}}
~\equiv~
\frac{\partial}{\partial \bm{W}}
\Bigl[\,
\tfrac{\bm{\Delta}}{2}~\Bigl(\lfloor x\rceil +0.5\Bigr)\Bigr].
\]
Noting that \(\tfrac{\bm{\Delta}}{2}\) does not depend on \(\bm{W}\), 
we mainly examine \(\frac{\partial}{\partial \bm{W}}\lfloor x\rceil\).  
In Quantization-Aware Training (QAT), the Straight-Through Estimator (STE) approximates:
\[
\frac{\partial}{\partial x}\bigl(\lfloor x\rceil\bigr)
~\approx~
1
\quad(\text{except at integer boundaries}).
\]
Hence, effectively, \(\lfloor x\rceil \approx x\) in backprop.
\paragraph{Clipping impact.}
Recall \(x=2\,z-0.5\) and \(z=\text{clip}\bigl(\tfrac{\bm{W}}{\bm{\Delta}}, -1+\epsilon, 1-\epsilon\bigr)\).  
If \(\lvert\,\tfrac{W_{ij}}{\Delta_{i}}\rvert>1-\epsilon\), then \(z_{ij}\) saturates to \(\pm(1-\epsilon)\) and its derivative \(\tfrac{\partial z_{ij}}{\partial W_{ij}}=0\).  
Otherwise, \(\tfrac{\partial z_{ij}}{\partial W_{ij}}=\tfrac{1}{\Delta_{i}}\).  
Since \(x=2\,z-0.5\), we get 
\(\tfrac{\partial x_{ij}}{\partial W_{ij}}=2\times\tfrac{\partial z_{ij}}{\partial W_{ij}}=\tfrac{2}{\Delta_{i}}\) in the non-saturated zone, or 0 if saturated.
\paragraph{Resulting piecewise gradient.}
Putting these together:
\begin{align*}
\frac{\partial\,\bm{W}_{\text{FP16}\rightarrow\text{INT2}}}{\partial\,\bm{W}}
&~\approx~
\tfrac{\bm{\Delta}}{2}\,
\underbrace{\Bigl(\tfrac{\partial \lfloor x\rceil}{\partial x}\Bigr)}_{\approx1}
\underbrace{\Bigl(\tfrac{\partial x}{\partial \bm{W}}\Bigr)}_{\text{0 or } \tfrac{2}{\Delta}}
\\
&=\;
\begin{cases}
\tfrac{\bm{\Delta}}{2}\times 1 \times \tfrac{2}{\bm{\Delta}}~=~1,
& \text{if }\,\bigl|\tfrac{W_{ij}}{\Delta_{i}}\bigr|\le 1-\epsilon,\\[6pt]
0, 
& \text{otherwise (saturated)}.
\end{cases}
\end{align*}
Therefore,
\[
\frac{\partial\,\bm{W}_{\text{FP16}\rightarrow\text{INT2}}}{\partial\,\bm{W}}
~\approx~
\begin{cases}
1,& \lvert W/\Delta\rvert \le 1-\epsilon,\\
0,& \lvert W/\Delta\rvert > 1-\epsilon.
\end{cases}
\]
\subsection{Gradient with respect to Scale}
Now we turn to 
\(\frac{\partial}{\partial \bm{\Delta}}\,\bm{W}_{\text{FP16}\rightarrow\text{INT2}}\).  
Again, from \eqref{eq:original_seq},
\[
\bm{W}_{\text{FP16}\rightarrow\text{INT2}}
~=~
\tfrac{\bm{\Delta}}{2}\,\bigl(\lfloor x\rceil +0.5\bigr),
\]
\paragraph{Decomposing the derivative.}
\[
\frac{\partial \bm{W}_{\text{FP16}\rightarrow\text{INT2}}}{\partial \bm{\Delta}}
~=~
\underbrace{\frac{\partial}{\partial\bm{\Delta}}
\Bigl(\tfrac{\bm{\Delta}}{2}\Bigr)}_{=\tfrac12}~
\bigl(\lfloor x\rceil+0.5\bigr)
~
+~
\tfrac{\bm{\Delta}}{2}
~
\underbrace{\frac{\partial \lfloor x\rceil}{\partial x}}_{\approx 1}
~
\underbrace{\frac{\partial x}{\partial \bm{\Delta}}}_{\text{clip-based}}.
\]
Hence:
\[
\frac{\partial \bm{W}_{\text{FP16}\rightarrow\text{INT2}}}{\partial \bm{\Delta}}
~\approx~
\tfrac12\,\lfloor x\rceil
~
+~
\tfrac{\bm{\Delta}}{2}\cdot 1 \cdot \tfrac{\partial x}{\partial \bm{\Delta}}.
\]
\paragraph{Clip-based partial of \(x\).}
Recall \(x=2\cdot\text{clip}\bigl(\tfrac{\bm{W}}{\bm{\Delta}},-1+\epsilon,1-\epsilon\bigr)-0.5\).  
In the non-saturated zone, \(\text{clip}(u)=u\), so 
\(\tfrac{\partial}{\partial \bm{\Delta}}\bigl(\tfrac{W_{ij}}{\Delta_i}\bigr)=-\tfrac{W_{ij}}{\Delta_i^2}\).  
Thus,
\[
\tfrac{\partial x_{ij}}{\partial \Delta_i}
~=~
2
\Bigl(-\frac{W_{ij}}{\Delta_i^2}\Bigr)
~=~
-2\,\frac{W_{ij}}{\Delta_i^2},
\quad
\text{if}\,
\bigl|\tfrac{W_{ij}}{\Delta_i}\bigr|\le 1-\epsilon,
\]
and 0 otherwise.
\paragraph{Putting it all together (piecewise).}
From this, we get results as follows:
\[
\frac{\partial \bm{W}_{\text{FP16}\rightarrow\text{INT2}}}{\partial \bm{\Delta}}
~=~
\begin{cases}
\dfrac{\bm{W}_{\text{FP16}\rightarrow\text{INT2}}}{\bm{\Delta}},
& \text{(if saturated, i.e.\ }\lvert W/\Delta\rvert > 1-\epsilon),\\[6pt]
\dfrac{\bm{W}_{\text{FP16}\rightarrow\text{INT2}} - \bm{W}}{\bm{\Delta}},
& \text{(if unsaturated, i.e.\ }\lvert W/\Delta\rvert \le 1-\epsilon).
\end{cases}
\]

Summarizing the findings, saturated weights (mapped to \(\pm3\)) completely lose their update signal with respect to \(\bm{W}\) (gradient=0), since further changes in \(\bm{W}\) do not alter the quantized value in that range. Conversely, those same saturated weights yield a strong gradient signal for \(\bm{\Delta}\).  If \(\lvert w_q\rvert=1.5\,\Delta\), then \(\tfrac{w_q}{\Delta}=\pm1.5\).  This can drive \(\Delta\) to adapt quickly, potentially pulling the weight back into the unsaturated zone (or saturating others further) depending on the loss objective. Hence, more saturated weights can imply less weight-level learning, but more \(\Delta\)-level learning.

Empirically, one might observe fewer weights in the \(\pm3\) bins if starting QAT directly from an FP checkpoint.  
This can be explained by the gradient formulas above:  
\begin{itemize}
\item In the unsaturated zone, the scale gradient is \(\tfrac{w_q - w}{\Delta}\). If \(w\approx w_q\) initially, this difference is small, so \(\Delta\) is not driven to expand or shrink aggressively.
\item With \(\Delta\) remaining relatively stable, fewer weights cross the \(\pm(1-\epsilon)\) boundary, so fewer get saturated.
\end{itemize}
On the other hand, starting from a PTQ-applied checkpoint might already scatter weights so that more lie near or beyond that boundary, thus yielding a higher fraction of \(\pm3\)-saturated weights and correspondingly larger scale gradients.

\newpage
\section{Limitations} \label{subsec:limitation}
While UPQ demonstrates the effectiveness of unified framework of progressive quantization for instruction-tuned LLMs, several directions remain open as unsolved problems for future works. First, our current framework primarily focuses on weight-only quantization, leaving activations in higher precision (e.g., FP16). Extending UPQ to include activation quantization would unlock the memory and latency benefits of extremely low-bit inference. Second, our experiments evaluate models up to moderate scales; examining whether UPQ generalizes consistently to much larger language models (e.g., 100B+ parameters) is an important question to answer. Third, although UPQ preserves a broad range of intrinsic capabilities, including instruction-following and reasoning skills, there may be domain-specific or multimodal tasks (e.g., code generation, image-text given reasoning) that would require additional fine-tuning techniques or specialized data. So, UPQ could potentially contribute to wider range of tasks. We leave these aspects as promising future works toward more comprehensive and effective low-bit instruction-tuned LLMs.

%%%%%%%%%%%%%%%%%%%%%%%%%%%%%%%%%%%%%%%%%%%%%%%%%%%%%%%%%%%%

% \newpage
% \input{9_checklist}

\end{document}